\documentclass{article}

\PassOptionsToPackage{numbers,sort&compress}{natbib}
\usepackage[preprint]{neurips_2026}

\usepackage[utf8]{inputenc} 
\usepackage[T1]{fontenc}    
\usepackage{hyperref}       
\usepackage{url}            
\usepackage{booktabs}       
\usepackage{amsfonts}       
\usepackage{nicefrac}       
\usepackage{microtype}      
\usepackage{xcolor}         
\usepackage{enumitem}       
\usepackage[normalem]{ulem} 

\usepackage{amsmath}
\usepackage{comment}
\usepackage{graphicx}
\usepackage{multirow}
\usepackage{subcaption}
\usepackage[capitalise]{cleveref}
\usepackage{tikz}
\usetikzlibrary{arrows.meta,positioning}
\usepackage{wrapfig}

\title{AsyncOPD: How Stale Can On-Policy Distillation Be?}

\author{
\parbox{\textwidth}{\centering
Wonjun Kang\thanks{Equal Contribution. Emails: \texttt{\{kangwj1995, kevin.galim\}@furiosa.ai}.}~~\textsuperscript{\mdseries 1}\quad
Kevin Galim$^{*}$\textsuperscript{\mdseries 1}\quad
Seunghyuk Oh\textsuperscript{\mdseries 1}\quad 
Minjun Kang\textsuperscript{\mdseries 2}\quad
Sanghyun Park\textsuperscript{\mdseries 2} \\
Donghoon Kim\textsuperscript{\mdseries 1}\quad
Minjae Lee\textsuperscript{\mdseries 1}\quad
Minseo Kim\textsuperscript{\mdseries 1}\quad
Rishabh Tiwari\textsuperscript{\mdseries 3}\quad
Yuchen Zeng\textsuperscript{\mdseries 4} \\
Hyung Il Koo\textsuperscript{\mdseries 1,2}\quad 
Kangwook Lee\textsuperscript{\mdseries 5,6} 
}
\\
\\
\parbox{\textwidth}{\centering
\textsuperscript{\mdseries 1} FuriosaAI \quad 
\textsuperscript{\mdseries 2} Ajou University \quad
\textsuperscript{\mdseries 3} UC Berkeley \quad
\textsuperscript{\mdseries 4} Microsoft Research \quad \\
\textsuperscript{\mdseries 5} KRAFTON \quad 
\textsuperscript{\mdseries 6} Ludo Robotics
}
\\
\\ 
Code: \url{https://github.com/furiosa-ai/async-opd}
}

\begin{document}

\maketitle

\begin{abstract}
On-policy distillation (OPD) trains a student on its own rollouts guided by teacher feedback and is becoming increasingly important for large language model (LLM) post-training. Like reinforcement learning (RL), however, OPD faces an on-policy systems bottleneck, as rollouts can dominate training time for reasoning workloads. Asynchronous training pipelines can alleviate this bottleneck by decoupling rollout generation from learner updates, but doing so introduces stale-policy data. While prior work has studied stale data in asynchronous RL, its effects in OPD remain underexplored. 
We present the first systematic study of staleness in asynchronous OPD, focusing on a practical setting where teacher feedback is implemented through local KL losses and full-vocabulary teacher logits are too expensive to store or transfer, necessitating finite teacher-score caches.
We first show that KL direction changes the stale-data problem: teacher-weighted forward KL is more robust to stale rollouts, whereas student-weighted reverse KL is vulnerable. Second, for this vulnerable reverse-KL case, we study whether methods designed to stabilize asynchronous RL can mitigate OPD staleness. In our experiments, they do not improve over a simpler OPD-specific surrogate: recomputing the reverse-KL signal under the current student at learner time. Third, we analyze how finite teacher-score caches create a bias-variance tradeoff for sparse and sampled reverse-KL OPD estimators. This motivates multi-sample Monte Carlo (MC), which preserves MC correctability while reducing one-sample variance. Finally, we present and open-source \textbf{AsyncOPD}, a fully asynchronous OPD training pipeline built from these estimator choices. Experiments show that AsyncOPD improves training throughput by $1.6\times$ to $3.8\times$ over strict synchronous training while reaching comparable accuracy.
\end{abstract}

\section{Introduction}
\label{sec:introduction}

\begin{figure}[!t]
\centering
\includegraphics[width=\linewidth]{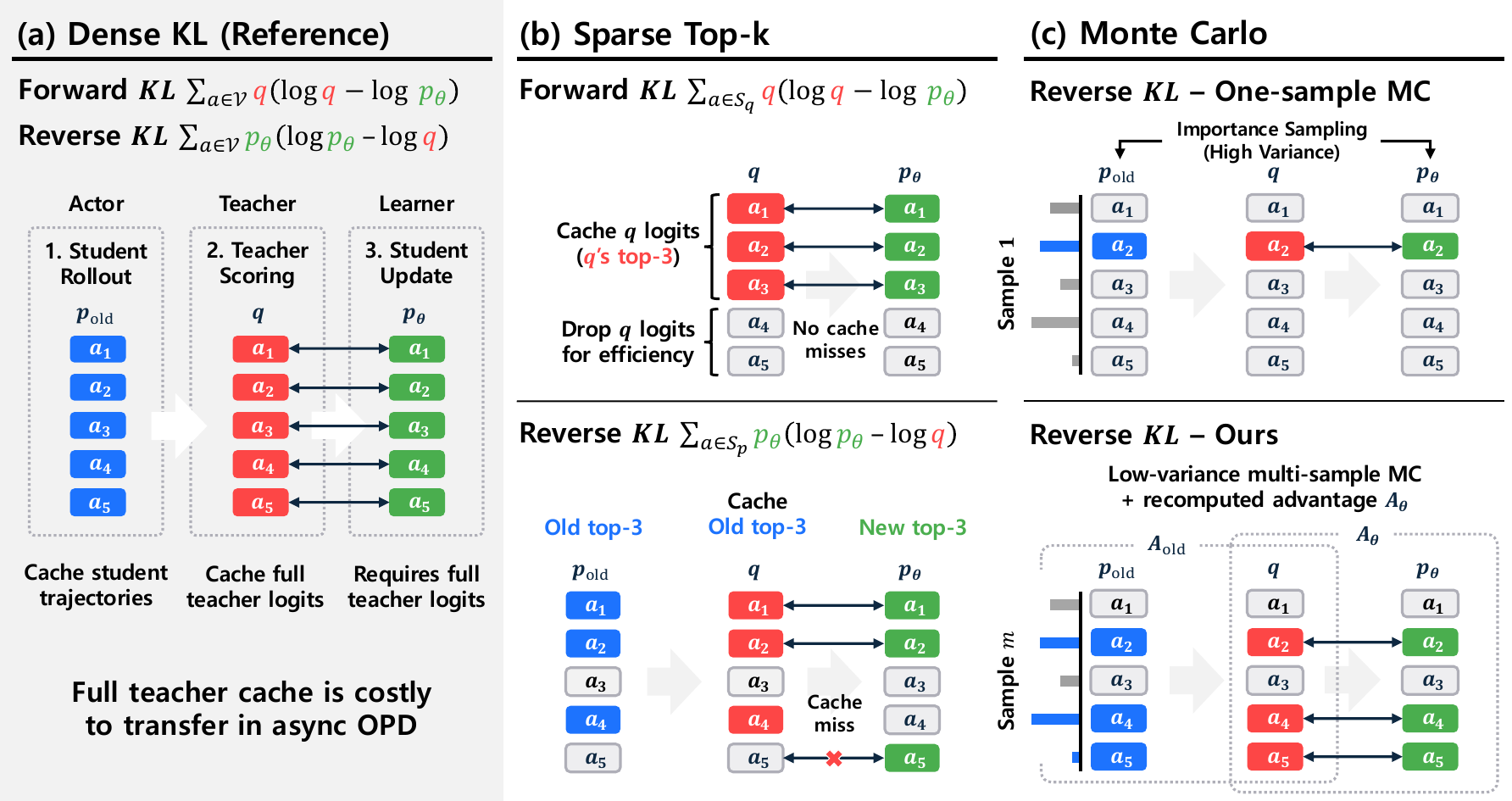}
\caption{Estimator design for asynchronous OPD.
(a) Dense KL is the full-vocabulary reference, but full teacher-logit caches are costly to store or transfer in asynchronous OPD.
(b) Sparse top-$k$ exposes a support mismatch under staleness: forward KL is teacher-supported, but reverse KL is student-supported and may require actions outside the cached teacher-scored support.
(c) One-sample Monte Carlo is correctable in expectation by importance sampling, but has high variance; our estimator recomputes $A_\theta$ at learner time and uses multi-sample MC to reduce variance.}
\label{fig:teaser}
\end{figure}

On-policy distillation
(OPD)~\citep{opd-survey,minillm,gkd} and reinforcement learning (RL)~\citep{rl-survey,dapo} have become central post-training methods
for improving large language model (LLM) reasoning~\citep{deepseek-r1},
including mathematics~\citep{grpo} and coding~\citep{glm-5}. 
OPD trains a student on its own rollouts using dense token-level feedback from a
teacher~\citep{tml-opd}, whereas RL learns from reward feedback on rollouts.
OPD provides an effective and efficient route for LLM post-training, especially
for smaller student models~\citep{qwen3}.
Recent work shows that OPD is not limited to distilling large teachers into
small students: it also supports on-policy self-distillation~\citep{opsd} and
multi-teacher distillation from domain-specialized teachers comparable in size
to the student~\citep{deepseekv4,mimo-v2-flash}.

OPD and RL inherit an on-policy systems bottleneck: each learner
update must wait for fresh rollouts from the model being trained~\citep{rollpacker}. For reasoning
tasks, these rollouts are long and expensive, so synchronous training often
waits on generation rather than updating the model, leaving learners
underutilized. 
Asynchronous RL~\citep{async-rlhf,areal} relieves this
bottleneck by decoupling rollout generation from learner updates: rollout
workers keep generating data while the learner updates on earlier rollouts,
improving training efficiency and hardware utilization~\citep{areal-hex,streamrl,laminar}.
A similar pipeline can be applied to OPD by running student rollout, teacher
scoring, and learner updates in parallel~\citep{verl}.

However, asynchronous execution introduces stale-policy data, and learning from such data can degrade model quality~\citep{scalerl}.
This creates a trade-off: more aggressive asynchrony improves training throughput, but it also increases the policy lag between rollout and learning.
Prior work on asynchronous RL
therefore studies how to stabilize learning from stale-policy
data~\citep{areal,m2po,a-3po}. However, it remains underexplored whether these
ideas and stale-data solutions transfer to OPD, because practical
implementations of OPD expose a different feedback interface. Teacher feedback
is often implemented through local KL losses, which require teacher scores over
actions at student-visited prefixes. Since full-vocabulary teacher logits are
expensive to store or transfer, especially in an asynchronous pipeline, teacher
scores are usually cached only on a finite set of actions~(\cref{fig:teaser}). Once the learner
receives a teacher-scored cache, it can recompute current-student log
probabilities on cached actions, but it cannot recover teacher scores for
actions that were never scored.
This raises three questions that structure
our study: (i) how asynchronous OPD behaves under staleness, (ii) whether
asynchronous RL ideas and stale-data solutions transfer to OPD, and (iii) how
finite teacher-score caches shape OPD estimator design. 

First, we study how KL direction shapes staleness. 
Under asynchronous OPD with cached teacher scores, the same stale rollout
cache can affect different KL objectives differently.
As illustrated in
\cref{fig:teaser}, forward KL is teacher-weighted and is more robust to stale
rollouts, whereas reverse KL is student-weighted and becomes vulnerable when
current-student actions fall outside the scored cache. We therefore focus on
reverse-KL OPD in the remainder of the staleness analysis.

Second, focusing on the reverse-KL case, we ask whether methods designed to
stabilize asynchronous RL can also mitigate OPD staleness. This comparison is
natural because reverse KL in OPD admits an RL-style policy-gradient surrogate, where
the teacher-student log-ratio acts as a token-level advantage. We therefore
evaluate PPO-style clipping~\citep{ppo}, decoupled PPO~\citep{areal}, and
M2PO~\citep{m2po}. In our experiments, they do not improve over a simpler
OPD-specific surrogate: recomputing the reverse-KL token-level advantage under
the current student at learner time without clipping.

Third, we return to the teacher-cache constraint and study the resulting
bias-variance tradeoff for sparse and sampled reverse-KL OPD implementations.
Stale student top-$k$ supports provide deterministic coverage but are
support-mismatched because they may omit actions required by the current
top-$k$ objective, and reweighting inside the stale support cannot recover the
missing teacher scores. One-sample Monte Carlo (MC) avoids this fixed-support
mismatch through importance-correctable samples from the stale rollout policy,
but suffers from high variance. This motivates multi-sample MC, which caches
and teacher-scores multiple stale-policy samples at each decoding step,
preserving MC correctability while reducing one-sample variance.

Finally, we instantiate these findings in \textbf{AsyncOPD}, a fully
asynchronous OPD pipeline that overlaps student rollout, teacher scoring, and
learner updates. On Qwen3-Base models, AsyncOPD improves training throughput
by $1.6\times$ to $3.8\times$ over strict synchronous training while
maintaining comparable accuracy.
Our contributions are:
\begin{itemize}[leftmargin=*]
  \item We provide the first systematic study of staleness in asynchronous OPD through the lens of an OPD-specific teacher-cache constraint.
  \item We show that KL direction changes the stale-data problem: forward KL is comparatively robust to stale rollouts, whereas reverse KL is vulnerable because it is
  student-weighted~(\cref{sec:opd-staleness}).
  \item We identify that the most effective reverse-KL policy-gradient surrogate uses the advantage recomputed at learner time without clipping, and that
  advanced asynchronous RL surrogates do not improve over this
  choice~(\cref{sec:reverse-kl-estimator-design}).
  \item We show that stale student top-$k$ supports are support-mismatched,
  while one-sample MC remains correctable but high-variance; this motivates
  multi-sample MC~(\cref{sec:estimator-design}).
  \item We present and open-source \textbf{AsyncOPD}, a fully asynchronous OPD
  training pipeline, and demonstrate improved
  training efficiency while maintaining OPD quality~(\cref{sec:asyncopd}).
\end{itemize}

\section{Related Works}

\paragraph{On-Policy Distillation}

On-policy distillation (OPD) trains a student on its own rollouts while using a
teacher to provide dense token-level feedback on the visited
prefixes~\citep{opd-survey,tml-opd}.
GKD~\citep{gkd} introduced a token-level KL
formulation, while MiniLLM~\citep{minillm} studied a
sequence-level reverse-KL variant. \citet{li2026rethinking} study token-level
OPD training dynamics and recipes for unstable configurations. TIP~\citep{tip}
characterizes per-token importance through student entropy and teacher-student
divergence. G-OPD~\citep{g-opd} interprets token-level OPD as dense
KL-constrained RL and extends it with reward scaling. 
These works clarify OPD as an effective post-training objective, but assume
rollouts, teacher scoring, and learner updates stay synchronized.

\paragraph{Asynchronous RL}

In synchronous RL pipelines, training often waits for the longest rollout in a
batch to finish, leaving learner resources idle. Asynchronous RL improves
hardware utilization by decoupling rollout generation from learner updates.
Async RLHF~\citep{async-rlhf} overlaps generation and learning so that new
samples are produced while the learner trains on earlier ones. StreamRL~\citep{
streamrl} further disaggregates the RLHF pipeline into streaming stages.
AReaL~\citep{areal} fully decouples rollout workers from training workers for
continuous asynchronous execution. Laminar~\citep{laminar} uses
fine-grained weight synchronization for trajectory-level asynchrony. However,
asynchronous RL must learn from stale-policy data. Decoupled PPO~\citep{areal}
stabilizes asynchronous RL training by separating the behavior policy for stale
rollouts from the proximal policy that anchors PPO~\citep{ppo} updates. M2PO~\citep{m2po}
stabilizes stale updates with second-moment importance-weight constraints, and
A-3PO~\citep{a-3po} reduces decoupled PPO overhead through staleness-aware
interpolation.

\paragraph{Asynchronous OPD}

VeRL~\citep{verl} implements step-off OPD schedulers that overlap student
rollout, teacher scoring, and learner update by fixing rollout lag to one or two
learner steps. These schedulers establish the practical feasibility of
asynchronous OPD, but leave open how OPD estimators behave under stale
teacher-scored caches. KDFlow~\citep{kdflow} improves systems efficiency for LLM
distillation by decoupling teacher inference from learner training and
transmitting teacher hidden states, but targets synchronous OPD and leaves
asynchronous execution as future work. We study this missing asynchronous OPD
regime directly and build AsyncOPD from the resulting estimator choices.

\section{Preliminaries: On-Policy Distillation}
\label{sec:preliminaries}

\paragraph{OPD setup}
At each decoding timestep, we view the visited prefix $s$ as the local state
and the next token $a$ as the action. Let $q(a \mid s)$ denote the teacher
policy and $p_\theta(a \mid s)$ denote the current student policy.
Following prior work on token-level OPD~\citep{tml-opd,g-opd,li2026rethinking},
we apply local losses to generated output tokens and analyze the resulting
objectives at a fixed prefix state $s$.
OPD can be defined with different divergences; forward and reverse KL are two
standard choices~\citep{gkd}.

\paragraph{Forward-KL OPD}

At a fixed prefix $s$, forward-KL OPD is teacher-weighted:
\begin{align}
D_F(\theta; s)
&=
\mathrm{KL}\!\left(q(\cdot \mid s)\,\|\,p_\theta(\cdot \mid s)\right)
=
\textstyle\sum\nolimits_{a\in\mathcal{V}} q(a\mid s)
\left(\log q(a\mid s)-\log p_\theta(a\mid s)\right).
\label{eq:dense-forward-kl}
\end{align}
At a fixed prefix $s$, the gradient is $\nabla_\theta D_F(\theta; s) = -\textstyle\sum\nolimits_{a\in\mathcal{V}}q(a\mid s)\nabla_\theta\log p_\theta(a\mid s).$

\paragraph{Reverse-KL OPD}

At the same prefix, reverse-KL OPD is student-weighted:
\begin{align}
D_R(\theta; s)
&=
\mathrm{KL}\!\left(p_\theta(\cdot \mid s)\,\|\,q(\cdot \mid s)\right)
=
-\textstyle\sum\nolimits_{a\in\mathcal{V}}p_\theta(a\mid s)
\left(\log q(a\mid s)-\log p_\theta(a\mid s)\right).
\label{eq:dense-reverse-kl}
\end{align}
Differentiating and using
$\mathbb{E}_{a\sim p_\theta(\cdot\mid s)}
[\nabla_\theta \log p_\theta(a\mid s)]
=\nabla_\theta\sum_a p_\theta(a\mid s)=0$ gives
\begin{align}
\nabla_\theta D_R(\theta; s)
&=
\textstyle\sum\nolimits_{a\in\mathcal{V}}
p_\theta(a\mid s)
\left(
\log p_\theta(a\mid s)-\log q(a\mid s)+1
\right)
\nabla_\theta\log p_\theta(a\mid s) \nonumber\\
&=
-\textstyle\sum\nolimits_{a\in\mathcal{V}}
p_\theta(a\mid s)
\left(\log q(a\mid s)-\log p_\theta(a\mid s)\right)
\nabla_\theta\log p_\theta(a\mid s).
\label{eq:reverse-kl-gradient}
\end{align}
Viewing \cref{eq:reverse-kl-gradient} as a policy-gradient estimator and
$A=\log q(a\mid s)-\log p(a\mid s)$ as the advantage term connects reverse-KL
OPD to standard RL training machinery, and practical implementations typically
use PPO-style surrogates. Given behavior-policy samples
$a\sim p_{\mathrm{beh}}$, define
$\rho_\theta(a,s)=p_\theta(a\mid s)/p_{\mathrm{beh}}(a\mid s)$ and
$\bar\rho_\theta(a,s)=\operatorname{clip}(\rho_\theta(a,s),1-\epsilon,1+\epsilon)$.
The PPO-style local surrogate uses these ratios with a frozen behavior-time
signal $A_{\mathrm{beh}}(a,s)$, where $\operatorname{sg}(\cdot)$ denotes stop-gradient:
\begin{align}
L_{\mathrm{PPO}}(\theta; A_{\mathrm{beh}})
=
-\mathbb{E}_{a\sim p_{\mathrm{beh}}}
\left[
\min\left(
\rho_\theta\operatorname{sg}\!\left(A_{\mathrm{beh}}\right),
\bar\rho_\theta\operatorname{sg}\!\left(A_{\mathrm{beh}}\right)
\right)
\right].
\label{eq:ppo-style-surrogate}
\end{align}
\paragraph{Sparse and sampled implementations}

The dense objectives above are full-vocabulary references. Practical OPD
instead evaluates local KL losses on finite supports or sampled
actions~\citep{tml-opd,li2026rethinking}, trading computation against support coverage and estimator variance.
\textbf{Sparse top-$k$} implementations choose a support $S(s)$ and evaluate the
corresponding restricted KL after renormalizing teacher and student
distributions on $S(s)$. \textbf{Monte Carlo (MC)} implementations draw actions from
a proposal distribution and estimate the corresponding local gradient; for
reverse KL, this yields the
student-sampled policy-gradient estimator. Details are in
\cref{app:sparse-implementations}.


\providecommand{\experimenttakeaway}[2]{%
\begin{center}
\fbox{\begin{minipage}{0.96\linewidth}
\textbf{Finding #1.} #2
\end{minipage}}
\end{center}
}

\section{Forward- and Reverse-KL OPD Under Staleness}
\label{sec:opd-staleness}

Asynchronous OPD has both prefix-level and action-level staleness. Once a
rollout is generated, its visited prefixes are fixed, so an action-level
estimator cannot change which states the learner sees. We therefore focus on
the action-level staleness that estimator design can directly address.

\subsection{Asynchronous OPD Setup}

Asynchronous OPD is a cached-data pipeline: rollout first selects prefixes and
actions, teacher scoring then annotates those actions, and the learner updates
the student later. Unlike synchronous OPD, these stages are separated in time,
so the visited prefixes, action cache, teacher scores, and update policy may be
tied to different student versions.
\Cref{fig:teaser} summarizes this cached-teacher setting and the estimator
contrasts induced by the three-stage cache pipeline.

\paragraph{\textit{Teacher-cache constraint}}
Full-vocabulary teacher logits allow dense KL computation, but caching and transferring them is prohibitively expensive, especially in an asynchronous pipeline.
We therefore focus on sparse top-$k$ supports and MC
samples as the sparse and sampled cases.

\paragraph{Stage 1: Student rollout}
A rollout actor samples trajectories from a stale student $p_{\text{old}}$,
which fixes the visited prefixes $s$. At each prefix, it stores cached actions
$C_{\text{old}}(s)$ together with their rollout-time log probabilities under
$p_{\text{old}}$, such as a sampled token or a top-$k$ support.

\paragraph{Stage 2: Teacher scoring}
Let $C_{\text{score}}(s)$ denote the teacher-scored cache; it may come
from the rollout cache $C_{\text{old}}(s)$ or be selected by the teacher at
scoring time. Once teacher scoring is complete, teacher logits are available
only on $C_{\text{score}}(s)$.

\paragraph{Stage 3: Student update}
By learner update time, the student has moved to the current policy $p_\theta$.
The learner can recompute $\log p_\theta(a\mid s)$ for
$a\in C_{\text{score}}(s)$, but the current local OPD objective may place mass
on actions outside this teacher-scored cache, such as the current student top-$k$
support or current student-sampled actions. Thus the learner can update current
student probabilities on cached actions, but cannot recover teacher signals for
missing actions without additional teacher access.


\paragraph{Experimental setup}
Unless otherwise stated, we train a Qwen3-4B-Base student using a
Qwen3-30B-A3B-Instruct-2507 teacher~\citep{qwen3}. The training data is DeepMath~\citep{he2025deepmath}, filtered to 57{,}630 math problems with difficulty level at least 6, and we report final-checkpoint Avg@32 accuracy on
AIME24~\citep{aime2024}, AIME25~\citep{aime2025}, and AMC~\citep{amc}.
Experimental details are provided in
\cref{app:experimental-details}; dataset and metric details are in
\cref{app:datasets-metrics}.

\subsection{Forward KL vs. Reverse KL Under Staleness}

The KL direction fixes the action weighting: forward KL weights actions by the
teacher $q$, whereas reverse KL weights them by the student $p_\theta$.
With cached teacher scores, this weighting difference becomes a support-ownership difference (\cref{fig:teaser}).
Under a scored-cache restriction, this makes forward KL less exposed to stale
student action choices: it does not need to convert stale student-sampled
actions into a current-student expectation. Reverse KL instead depends on
student-weighted action terms, so the same asynchronous cache creates a
different action-level staleness problem.

\paragraph{Experimental results}

\begin{figure}[t]
\centering
\begin{subfigure}[t]{0.24\linewidth}
\centering
\includegraphics[width=\linewidth]{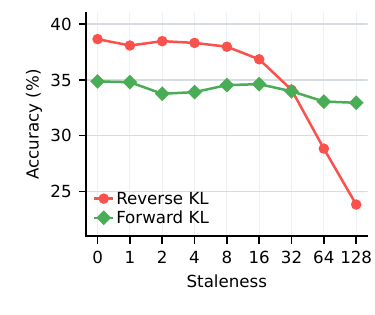}
\caption{Average}
\label{fig:mean-reverse-ppo-vs-forward-kl-final}
\end{subfigure}
\hfill
\begin{subfigure}[t]{0.24\linewidth}
\centering
\includegraphics[width=\linewidth]{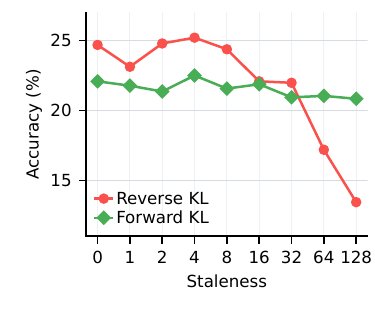}
\caption{AIME24}
\label{fig:aime24-reverse-ppo-vs-forward-kl-final}
\end{subfigure}
\hfill
\begin{subfigure}[t]{0.24\linewidth}
\centering
\includegraphics[width=\linewidth]{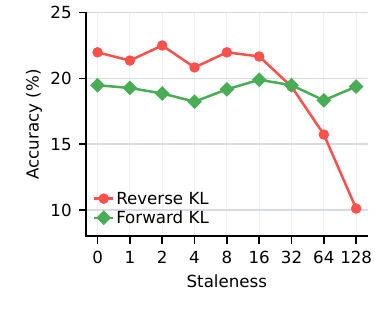}
\caption{AIME25}
\label{fig:aime25-reverse-ppo-vs-forward-kl-final}
\end{subfigure}
\hfill
\begin{subfigure}[t]{0.24\linewidth}
\centering
\includegraphics[width=\linewidth]{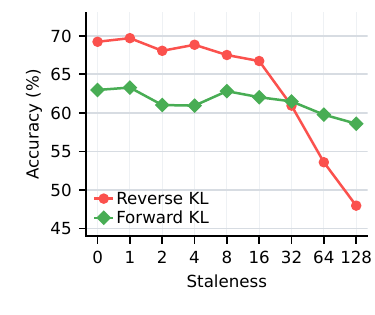}
\caption{AMC}
\label{fig:amc-reverse-ppo-vs-forward-kl-final}
\end{subfigure}
\caption{Accuracy comparison under staleness for forward- and reverse-KL OPD. Reverse KL starts higher at zero staleness but degrades faster as staleness grows; forward KL is flatter across the sweep.}
\label{fig:aime24-reverse-ppo-vs-forward-kl}
\end{figure}

\Cref{fig:aime24-reverse-ppo-vs-forward-kl} compares representative practical
OPD implementations from prior work: sparse top-$k$ forward KL
\citep{verl} and PPO-style reverse-KL surrogates \citep{tml-opd,g-opd}.
Reverse KL starts higher at
zero staleness, but as staleness increases it drops faster and is eventually
overtaken by forward KL. We therefore focus the rest of the staleness analysis on how to make reverse-KL
OPD robust under larger rollout staleness.

\experimenttakeaway{1}{Forward KL is teacher-weighted and robust to rollout
staleness, whereas reverse KL is student-weighted and vulnerable to rollout
staleness.}
\paragraph{Two axes of reverse-KL staleness}
The cache analysis above suggests a possible mechanism for this gap: because reverse KL is weighted by the current student, stale teacher-scored caches may fail to cover actions needed by the current reverse-KL objective.
In addition,
reverse-KL policy-gradient updates can be instantiated with multiple stale-data
surrogates, including PPO-style and asynchronous-RL variants.
We therefore split the reverse-KL
analysis into a \textbf{policy-gradient surrogate axis}, studied in
\cref{sec:reverse-kl-estimator-design}, and a \textbf{cached-support axis},
studied in \cref{sec:estimator-design}.

\section{Reverse-KL: Policy-Gradient Surrogates Under Staleness}
\label{sec:reverse-kl-estimator-design}

Reverse-KL OPD admits several policy-gradient surrogate choices under stale
rollouts. This section compares which choices remain effective under staleness.

\subsection{Policy-Gradient Surrogate Choices}
\label{sec:ppo-style-surrogates}

\paragraph{PPO-style objective}
In the PPO-style surrogate in \cref{eq:ppo-style-surrogate}, the advantage is
computed under the behavior policy and then held fixed during the learner
update. In stale reverse-KL OPD, the behavior policy is the rollout student, so a mechanical
PPO-style adaptation sets $p_{\mathrm{beh}}=p_{\mathrm{old}}$ and uses the
rollout-time reverse-KL advantage
$A_{\mathrm{old}}(a,s)=\log q(a\mid s)-\log p_{\mathrm{old}}(a\mid s)$
as $A_{\mathrm{beh}}$, together with the clipped old-to-current ratio. The
unclipped variant simply drops the clipped term.

\paragraph{Exact importance-sampling identity} 
In contrast, rewriting the reverse-KL objective~(\cref{eq:dense-reverse-kl}) by importance sampling suggests
a different surrogate choice. With the current reverse-KL advantage
$A_\theta(a,s)=\log q(a\mid s)-\log p_\theta(a\mid s)$, and assuming
$p_{\mathrm{old}}$ has support wherever $p_\theta$ does, the current reverse-KL
objective admits the exact old-to-current importance-sampling (IS) identity
\begin{align}
D_R(\theta;s)
=
-\mathbb{E}_{a\sim p_\theta}
\left[A_\theta(a,s)\right]
=
-\mathbb{E}_{a\sim p_{\mathrm{old}}}
\left[\rho_\theta(a,s)A_\theta(a,s)\right].
\end{align}
For the policy-gradient update, the advantage is used as a stop-gradient weight; the derivative of the omitted $A_\theta$ term cancels by the
score-function identity, as in \cref{eq:reverse-kl-gradient}.
Thus the IS view points to the opposite surrogate choice from the mechanical
PPO adaptation: recompute $A_\theta$ under the current student and use the
old-to-current ratio without clipping, with $A_\theta$ treated as a
stop-gradient advantage.

\paragraph{A two-by-two surrogate ablation}
\label{sec:local-signal}
\label{sec:local-signal-two-by-two}
The PPO-style adaptation and the OPD/IS identity suggest different surrogate
choices. We therefore ablate the advantage ($A_{\mathrm{old}}$ versus
$A_\theta$) and whether to clip the ratio, with $\operatorname{sg}(\cdot)$
denoting stop-gradient:
\begingroup
\small
\begin{align}
L_{\mathrm{old}}^{\mathrm{clip}}(\theta)
&=
-\mathbb{E}_{a\sim p_{\mathrm{old}}}
\left[
\min\left(
\rho_\theta\operatorname{sg}\!\left(A_{\mathrm{old}}\right),
\bar\rho_\theta\operatorname{sg}\!\left(A_{\mathrm{old}}\right)
\right)
\right],
&
L_{\mathrm{old}}^{\mathrm{noclip}}(\theta)
&=
-\mathbb{E}_{a\sim p_{\mathrm{old}}}
\left[
\rho_\theta\operatorname{sg}\!\left(A_{\mathrm{old}}\right)
\right],
\\
L_{\theta}^{\mathrm{clip}}(\theta)
&=
-\mathbb{E}_{a\sim p_{\mathrm{old}}}
\left[
\min\left(
\rho_\theta\operatorname{sg}\!\left(A_\theta\right),
\bar\rho_\theta\operatorname{sg}\!\left(A_\theta\right)
\right)
\right],
&
L_{\theta}^{\mathrm{noclip}}(\theta)
&=
-\mathbb{E}_{a\sim p_{\mathrm{old}}}
\left[
\rho_\theta\operatorname{sg}\!\left(A_\theta\right)
\right].
\label{eq:local-signal-theta-noclip}
\end{align}
\endgroup
Here $L_{\mathrm{old}}^{\mathrm{clip}}$ is the PPO-style adaptation, while
$L_{\theta}^{\mathrm{noclip}}$ is the OPD/IS surrogate.

\begin{figure}[!t]
\centering
\begin{subfigure}[t]{0.24\linewidth}
\centering
\includegraphics[width=\linewidth]{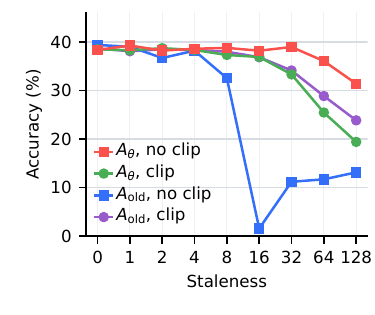}
\caption{Average}
\label{fig:mean-adv-clip-ablation-final}
\end{subfigure}
\hfill
\begin{subfigure}[t]{0.24\linewidth}
\centering
\includegraphics[width=\linewidth]{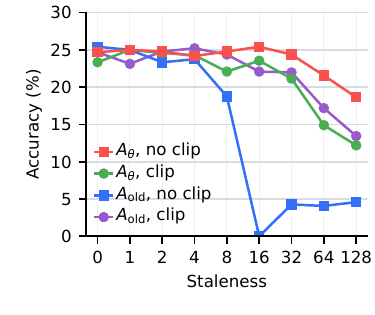}
\caption{AIME24}
\label{fig:aime24-adv-clip-ablation-final}
\end{subfigure}
\hfill
\begin{subfigure}[t]{0.24\linewidth}
\centering
\includegraphics[width=\linewidth]{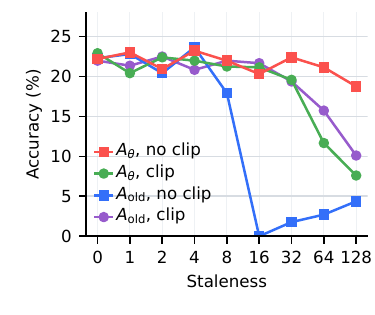}
\caption{AIME25}
\label{fig:aime25-adv-clip-ablation-final}
\end{subfigure}
\hfill
\begin{subfigure}[t]{0.24\linewidth}
\centering
\includegraphics[width=\linewidth]{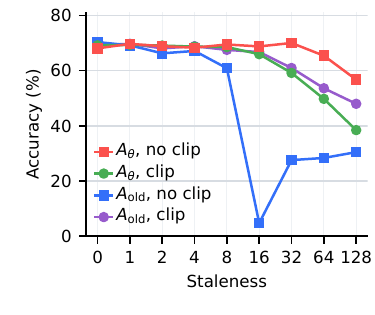}
\caption{AMC}
\label{fig:amc-adv-clip-ablation-final}
\end{subfigure}
\caption{Accuracy comparison under staleness for the advantage-and-clipping ablation. Recomputing $A_\theta$ at learner time and avoiding clipping gives the most stable performance across the sweep, while clipping mainly helps the frozen $A_{\mathrm{old}}$ baseline.}
\label{fig:aime24-adv-clip-ablation}
\end{figure}

\begin{figure}[!t]
\centering
\begin{subfigure}[t]{0.24\linewidth}
\centering
\includegraphics[width=\linewidth]{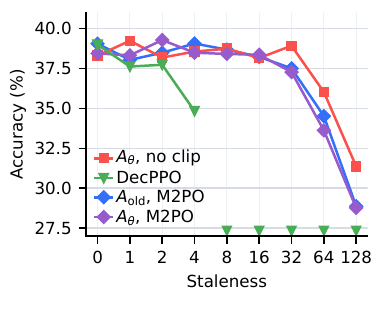}
\caption{Average}
\label{fig:mean-m2po-ablation-final}
\end{subfigure}
\hfill
\begin{subfigure}[t]{0.24\linewidth}
\centering
\includegraphics[width=\linewidth]{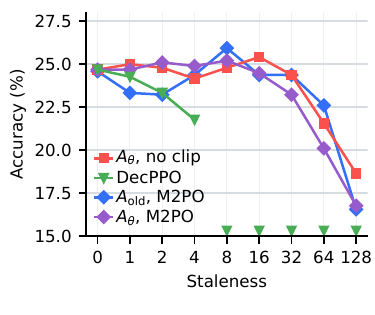}
\caption{AIME24}
\label{fig:aime24-m2po-ablation-final}
\end{subfigure}
\hfill
\begin{subfigure}[t]{0.24\linewidth}
\centering
\includegraphics[width=\linewidth]{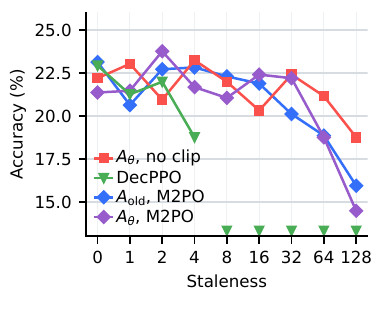}
\caption{AIME25}
\label{fig:aime25-m2po-ablation-final}
\end{subfigure}
\hfill
\begin{subfigure}[t]{0.24\linewidth}
\centering
\includegraphics[width=\linewidth]{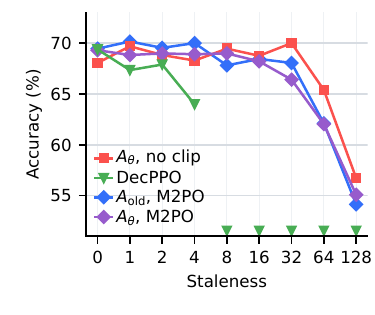}
\caption{AMC}
\label{fig:amc-m2po-ablation-final}
\end{subfigure}
\caption{Accuracy comparison under staleness for advanced asynchronous RL surrogates. Decoupled PPO~\citep{areal} and M2PO~\citep{m2po} do not consistently improve over the simpler OPD/IS surrogate that recomputes $A_\theta$ without clipping; Decoupled PPO is clipped for readability because of low accuracy.}
\label{fig:aime24-m2po-ablation}
\end{figure}

\paragraph{Advanced asynchronous RL surrogates}
\label{sec:async-rl-stabilizers}
Decoupled PPO~\citep{areal} and M2PO~\citep{m2po} are asynchronous RL surrogates
designed to improve robustness to stale-policy updates.
We evaluate whether these previously unstudied asynchronous RL surrogates also help OPD under staleness.

\begin{wrapfigure}{r}{0.25\linewidth}
\vspace{-32pt}
\centering
\includegraphics[width=\linewidth]{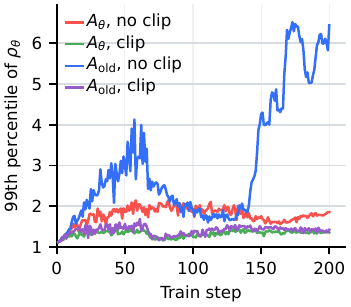}
\caption{$A_\theta$ reduces the p99 $\rho_\theta$ tail under no clip.}
\label{fig:r-p99-stepoff-so64}
\vspace{-10pt}
\end{wrapfigure}

\subsection{Experimental Results}
\label{sec:local-signal-experiments}

\cref{fig:aime24-adv-clip-ablation} and \cref{tab:local-signal-staleness-slopes} compare the four combinations of
$A_{\mathrm{old}}$ versus $A_\theta$ and clipping versus no clipping. The
best variant is the OPD/IS choice: $A_\theta$ without
clipping. The PPO-style baseline, $A_{\mathrm{old}}$ with clipping, remains a
strong stale-surrogate baseline.
Clipping helps $A_{\mathrm{old}}$ by limiting stale, large-ratio updates, but hurts $A_\theta$: recomputing $A_\theta$ already reduces the high-percentile $\rho_\theta$ tail at staleness 64 (\cref{fig:r-p99-stepoff-so64}), so clipping removes useful signal.
Likewise, \cref{fig:aime24-m2po-ablation} and
\cref{tab:local-signal-staleness-slopes} show that advanced asynchronous RL
surrogates such as decoupled PPO and M2PO do not outperform
$A_\theta$ without clipping, which becomes our reference surrogate below.

\experimenttakeaway{2}{The most effective reverse-KL correction is to recompute
$A_\theta$ at learner time without clipping; advanced asynchronous RL
surrogates such as decoupled PPO and M2PO do not improve over it.}

\begin{table}[!b]
\centering
\caption{Staleness-sensitivity slopes. Entries fit accuracy against
$\log_2(\mathrm{staleness}+1)$; more negative values indicate stronger
degradation with staleness.}
\label{tab:staleness-sensitivity-slopes}
\scriptsize
\setlength{\tabcolsep}{2.5pt}

\begin{subtable}[t]{0.65\linewidth}
\centering
\caption{Policy-gradient surrogates}
\label{tab:local-signal-staleness-slopes}
\resizebox{\linewidth}{!}{%
\begin{tabular}{@{}lrrrrrr@{}}
\toprule
Benchmark
& $A_{\mathrm{old}}$ (clip)
& $A_{\mathrm{old}}$
& $A_\theta$ (clip)
& $A_\theta$
& $A_{\mathrm{old}}$ (M2PO)
& $A_\theta$ (M2PO) \\
\cmidrule(r){1-1}\cmidrule(l){2-7}
AIME24 & -1.44 & -3.99 & -1.64 & \underline{-0.69} & \textbf{-0.66} & -1.00 \\
AIME25 & -1.42 & -3.69 & -1.88 & \textbf{-0.38} & \underline{-0.78} & -0.81 \\
AMC & -3.06 & -8.00 & -4.06 & \textbf{-1.12} & -1.75 & \underline{-1.72} \\
\bottomrule
\end{tabular}%
}
\end{subtable}
\hfill
\begin{subtable}[t]{0.33\linewidth}
\centering
\caption{Multi-sample MC}
\label{tab:cached-support-staleness-slopes}
\resizebox{\linewidth}{!}{%
\begin{tabular}{@{}lrrrr@{}}
\toprule
Benchmark
& MC1
& MC4
& MC16
& MC64 \\
\cmidrule(r){1-1}\cmidrule(l){2-5}
AIME24 & -0.69 & \underline{-0.42} & -0.46 & \textbf{-0.34} \\
AIME25 & -0.38 & \underline{-0.23} & -0.24 & \textbf{-0.11} \\
AMC & -1.12 & -0.87 & \textbf{-0.74} & \underline{-0.84} \\
\bottomrule
\end{tabular}%
}
\end{subtable}

\end{table}

\section{Reverse-KL: Cached Supports Under Staleness}
\label{sec:estimator-design}

Having fixed $A_\theta$ without clipping as the reference surrogate, we now ask
which cached actions provide the teacher scores needed to evaluate it, and how
to improve this cached-support estimator. This cached-support axis is specific
to OPD because teacher scoring is local and expensive: the teacher cache
determines which actions have teacher scores available to the learner.

\paragraph{Sparse top-\texorpdfstring{$k$}{k}: stale-support biased}
Although sparse top-$k$ is biased relative to the dense reverse-KL objective,
it is a practical low-variance approximation on the current student support
$S_\theta(s)=\operatorname{TopK}(p_\theta(\cdot\mid s),k)$. Under asynchronous
rollout reuse, however, teacher scores are cached on the rollout-time support
$S_{\mathrm{old}}(s)=\operatorname{TopK}(p_{\mathrm{old}}(\cdot\mid s),k)$,
which may miss actions in the current support $S_\theta(s)$. Reweighting within
$S_{\mathrm{old}}$ cannot recover these missing teacher scores, so stale sparse
top-$k$ remains a support-mismatched approximation, not an exact correction of
the current top-$k$ objective.

\paragraph{One-sample MC: correctable but high variance}
Sampled-token MC instead caches an action drawn from a behavior distribution:
$a\sim p_{\mathrm{old}}(\cdot\mid s)$ together with
$\log p_{\mathrm{old}}(a\mid s)$. When the behavior policy covers the current
policy support, exact old-to-current IS gives an unbiased estimator of the
current reverse-KL fixed-prefix gradient. Thus one-sample MC is action-level
correctable in expectation, but the resulting IS estimator can have high
variance. This proposal-sampling structure is the key contrast with stale
top-$k$, whose actions come from a deterministic stale support.

\subsection{Proposed Solution: Multi-Sample MC}

\paragraph{Multi-sample MC: correctable with reduced variance}
We propose multi-sample MC (\cref{fig:multisample-mc-schematic}), which, at each
decoding timestep of a student rollout, draws multiple local next-token samples
from the behavior policy without rolling them out into additional trajectories.
It reduces one-sample MC
variance by caching these local samples and averaging their IS-corrected
gradients.

Multi-sample MC is especially natural in asynchronous OPD.
In RL for LLM post-training, branching a prefix into multiple actions is expensive because each branch typically requires a full continuation before the reward or advantage can be evaluated.
In synchronous OPD, sparse top-$k$ already provides a low-variance
approximation and one-sample MC provides an unbiased sampled gradient
estimator, so there is little motivation to cache multiple sampled actions per
prefix.
Under asynchronous OPD, this tradeoff changes: sparse top-$k$ becomes the stale
fixed-support approximation analyzed above, and one-sample MC remains
correctable but high-variance, making multi-sample MC a natural cached-support
estimator for asynchronous OPD.
\begin{figure}[t]
\centering
\begin{subfigure}[t]{0.24\linewidth}
\centering
\includegraphics[width=\linewidth]{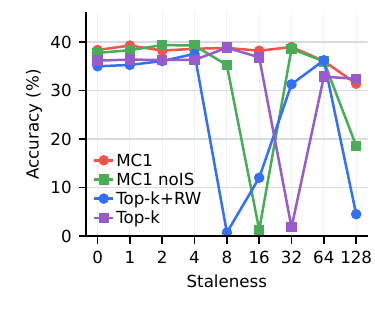}
\caption{Average}
\label{fig:mean-mc-vs-sparse-topk-final}
\end{subfigure}
\hfill
\begin{subfigure}[t]{0.24\linewidth}
\centering
\includegraphics[width=\linewidth]{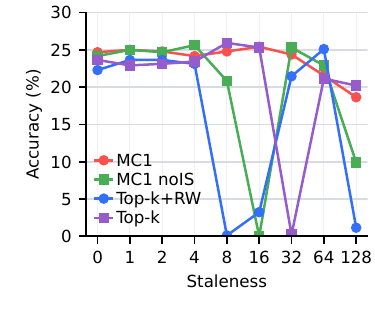}
\caption{AIME24}
\label{fig:aime24-mc-vs-sparse-topk-final}
\end{subfigure}
\hfill
\begin{subfigure}[t]{0.24\linewidth}
\centering
\includegraphics[width=\linewidth]{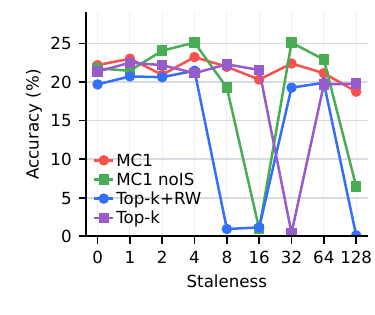}
\caption{AIME25}
\label{fig:aime25-mc-vs-sparse-topk-final}
\end{subfigure}
\hfill
\begin{subfigure}[t]{0.24\linewidth}
\centering
\includegraphics[width=\linewidth]{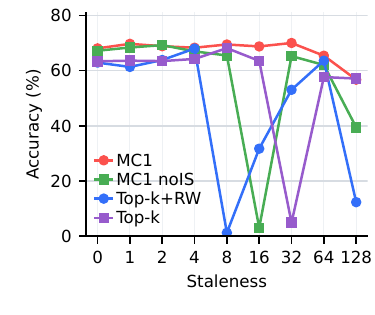}
\caption{AMC}
\label{fig:amc-mc-vs-sparse-topk-final}
\end{subfigure}
\caption{Accuracy comparison under staleness for sampled MC versus stale top-$k$. Top-$k$+RW denotes reweighting on the stale top-$k$ support. Old-to-current IS corrects sampled MC in expectation, whereas reweighting cannot repair the missing teacher scores induced by stale top-$k$ supports.}
\label{fig:aime24-mc-vs-sparse-topk}
\end{figure}

\begin{figure}[t]
\centering
\begin{subfigure}[t]{0.24\linewidth}
\centering
\includegraphics[width=\linewidth]{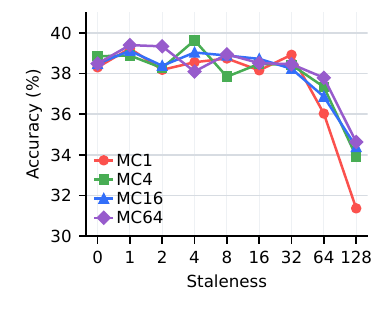}
\caption{Average}
\label{fig:mean-mc-sample-count-final}
\end{subfigure}
\hfill
\begin{subfigure}[t]{0.24\linewidth}
\centering
\includegraphics[width=\linewidth]{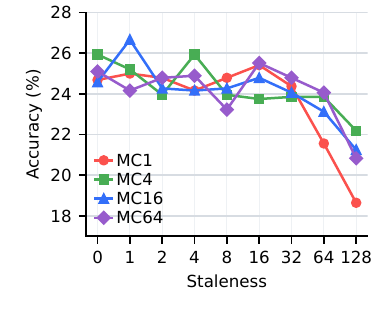}
\caption{AIME24}
\label{fig:aime24-mc-sample-count-final}
\end{subfigure}
\hfill
\begin{subfigure}[t]{0.24\linewidth}
\centering
\includegraphics[width=\linewidth]{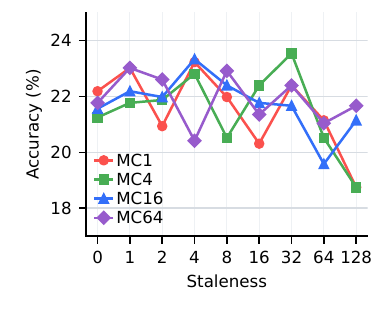}
\caption{AIME25}
\label{fig:aime25-mc-sample-count-final}
\end{subfigure}
\hfill
\begin{subfigure}[t]{0.24\linewidth}
\centering
\includegraphics[width=\linewidth]{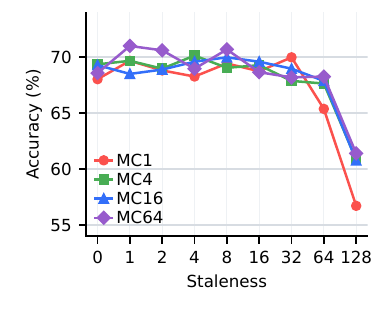}
\caption{AMC}
\label{fig:amc-mc-sample-count-final}
\end{subfigure}
\caption{Accuracy comparison under staleness for multi-sample MC. Increasing the number of samples improves large-staleness behavior.}
\label{fig:aime24-mc-sample-count}
\end{figure}

\begin{wrapfigure}{r}{0.37\linewidth}
\vspace{0pt}
\centering
\resizebox{\linewidth}{!}{%
\begin{tikzpicture}[
  font=\scriptsize,
  >=Latex,
  state/.style={draw=black!65, fill=black!4, rounded corners=2pt,
    text width=1.05cm, align=center, minimum height=0.52cm, inner sep=1.4pt},
  action/.style={draw=black!70, fill=black!6, rounded corners=2pt,
    text width=0.82cm, align=center, minimum height=0.52cm, inner sep=1.1pt},
  main/.style={-{Latex[length=1.6mm]}, thick, draw=black!82},
  branch/.style={draw=black!70, thick},
  guide/.style={draw=black!70, thick, densely dashed},
  op/.style={draw=black!70, fill=white, circle, inner sep=0pt,
    minimum size=0.10cm, font=\tiny},
  note/.style={font=\scriptsize, align=center},
  tiny/.style={font=\scriptsize}
]

\filldraw[draw=black!20, fill=black!3, rounded corners=2pt]
  (-0.65,-0.4) rectangle (5.44,0.9);

\node[font=\small, align=center] at (2.30,0.56) {one student rollout path};

\node[state] (s0) at (0,0) {\shortstack{$s_t$\\[-1pt]\tiny\texttt{hello}}};
\node[state] (s1) at (1.86,0) {\shortstack{$s_{t+1}$\\[-1pt]\tiny\texttt{... what}}};
\node[state] (s2) at (3.72,0) {\shortstack{$s_{t+2}$\\[-1pt]\tiny\texttt{... is}}};

\draw[main] (s0) -- (s1);
\draw[main] (s1) -- (s2);
\node[op] (op0) at (0.89,0) {$+$};
\node[op] (op1) at (2.75,0) {$+$};
\draw[main] (s2.east) -- ++(0.58,0);
\node[anchor=west] at (4.81,0) {$\cdots$};

\foreach \x/\stateNode in {0/s0,1.86/s1,3.72/s2} {
  \draw[branch] (\stateNode.south) -- (\x,-1.54);
}

\node[action] (a01) at (0.89,-0.82)
  {\shortstack{$a_{t,1}$\\[-1pt]\tiny\texttt{what}}};
\node[action] (a02) at (0.89,-1.54)
  {\shortstack{$a_{t,2}$\\[-1pt]\tiny\texttt{how}}};

\node[action] (a11) at (2.75,-0.82)
  {\shortstack{$a_{t+1,1}$\\[-1pt]\tiny\texttt{is}}};
\node[action] (a12) at (2.75,-1.54)
  {\shortstack{$a_{t+1,2}$\\[-1pt]\tiny\texttt{did}}};

\draw[guide] (a01.north) -- (op0.south);
\draw[guide] (a11.north) -- (op1.south);

\node[action] (a21) at (4.61,-0.82)
  {\shortstack{$a_{t+2,1}$\\[-1pt]\tiny\texttt{your}}};
\node[action] (a22) at (4.61,-1.54)
  {\shortstack{$a_{t+2,2}$\\[-1pt]\tiny\texttt{going}}};

\draw[branch] (0,-0.82) -- (a01.west);
\draw[branch] (0,-1.54) -- (a02.west);
\draw[branch] (1.86,-0.82) -- (a11.west);
\draw[branch] (1.86,-1.54) -- (a12.west);
\draw[branch] (3.72,-0.82) -- (a21.west);
\draw[branch] (3.72,-1.54) -- (a22.west);

\node[font=\small, align=center] at (2.30,-2.3)
  {example: $m=2$ local samples per timestep};

\end{tikzpicture}%
}
\caption{Multi-sample MC ($m=2$).}
\label{fig:multisample-mc-schematic}
\vspace{0pt}
\end{wrapfigure}

Concretely, at each visited timestep $t$ with prefix $s_t$, rollout samples
$a_{t,1},\ldots,a_{t,m}\sim p_{\mathrm{old}}(\cdot\mid s_t)$ and caches their
rollout log probabilities and teacher scores. For notational simplicity, write
$s=s_t$ and $a_i=a_{t,i}$ below. At learner time, we recompute $A_\theta(a_i,s)$
and use the averaged unclipped old-to-current IS surrogate
$\widehat L_m^{\mathrm{MC}}(\theta;s)=
-\frac{1}{m}\sum\nolimits_{i=1}^m
\rho_\theta(a_i,s)\operatorname{sg}(A_\theta(a_i,s))$. By linearity, the
gradient has the same expectation as the one-sample MC estimator; averaging
independent behavior-policy samples reduces the Monte Carlo variance. We measure
this variance reduction at large staleness in \cref{app:mc-variance-large-staleness}.

\subsection{Experimental Results}
\label{sec:cached-support-experiments}

\paragraph{Sparse top-$k$ vs. one-sample MC}
\Cref{fig:aime24-mc-vs-sparse-topk} compares one-sample MC and sparse
top-$k$, with and without old-to-current reweighting. For one-sample MC, IS
substantially improves robustness as staleness increases. For sparse top-$k$,
the same reweighting does not improve performance, since it cannot recover
missing current-support actions. As a result, one-sample MC with IS is the
strongest of the four methods, consistent with the support-correctability
analysis above. We include an additional ablation disentangling MC sample count
from IS in \cref{app:is-ablation}.

\paragraph{One-sample MC vs. multi-sample MC}
\Cref{fig:aime24-mc-sample-count} and
\cref{tab:cached-support-staleness-slopes} show that multi-sample MC improves
one-sample MC at large staleness: $m=4$ already gives a clear jump,
while $m\in\{4,16,64\}$ performs similarly.

\experimenttakeaway{3}{One-sample MC is more effective than stale sparse top-$k$; multi-sample MC further improves this
estimator by reducing one-sample variance while preserving MC correctability.}

\section{AsyncOPD: Fully Asynchronous OPD}
\label{sec:asyncopd}

AsyncOPD is our fully
asynchronous OPD system. Following AReaL~\citep{areal}, it overlaps rollout,
teacher scoring, and learner updates.



\begin{figure}[!t]
\centering
\includegraphics[width=\linewidth]{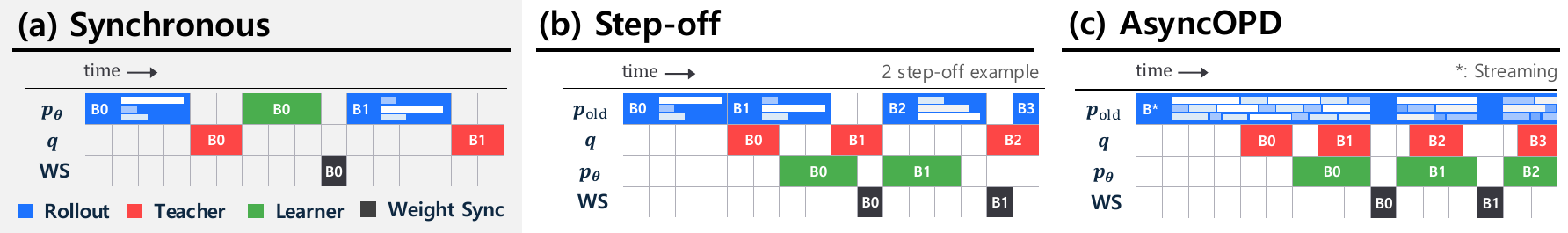}
\caption{Scheduler comparison for synchronous OPD, step-off scheduling, and AsyncOPD.
Synchronous OPD is barriered; step-off scheduling~\citep{verl} overlaps stages but keeps gated rollout batches, while AsyncOPD streams rollout data to reduce long-tail waiting.}
\label{fig:asyncopd-scheduler-timeline}
\end{figure}

\paragraph{Scheduler}
The step-off scheduler family was originally implemented in
VeRL~\citep{verl}: a $k$-step-off run fixes rollout lag to $k$ learner updates,
but still waits for complete rollout batches. AsyncOPD streams examples
instead: workers pause only for weight sync, preserve in-flight prefixes,
teacher scoring consumes completed items, and the learner updates once a scored
batch is ready
(\cref{fig:asyncopd-scheduler-timeline}).

\paragraph{Experimental setup}

The main comparison uses Qwen3-\{1.7B,4B,8B\}-Base students with the Qwen3-30B-A3B-Instruct-2507 teacher. All runs use the same reverse-KL estimator: current-policy $A_\theta$, no clipping, old-to-current IS, and either MC64 or MC1. We compare strict sync, two-step-off, and AsyncOPD for 100 training iterations on the same 8-GPU node; all AsyncOPD runs use $\tau=4$. \cref{app:asyncopd-scheduler-details} gives GPU allocation, queue-depth, and scheduler details.

\paragraph{Experimental Results}

\Cref{tab:asyncopd-scheduler-results} reports training throughput, pipeline overlap 
(average concurrent OPD-stage activity), and final AIME24 Avg@32 for the 
Qwen3-Base students. AsyncOPD achieves the highest throughput and overlap 
in every matched comparison. In MC64, it reaches up to $2.7\times$ the 
strict-sync throughput while achieving the best or tied-best final accuracy. 
MC1 shows the same trend: AsyncOPD delivers the highest throughput (up to $3.3\times$ strict-sync)
and overlap for every student, with competitive final accuracy.
\begin{table}[t]
\centering
\footnotesize
\setlength{\tabcolsep}{3pt}
\renewcommand{\arraystretch}{1}
\captionsetup{skip=3pt}
\caption{AsyncOPD scheduler results for Qwen3-Base models. Train tok/s is training throughput; parentheses show speedup over the matched strict-sync baseline. Overlap is concurrent OPD-stage activity. Avg@32 is final AIME24. AsyncOPD achieves the highest throughput and overlap in all matched settings while maintaining comparable final accuracy.}
\label{tab:asyncopd-scheduler-results}
\resizebox{\linewidth}{!}{%
\begin{tabular}{@{}llrrrrrr@{}}
\toprule
\multirow{2.5}{*}{Student} & \multirow{2.5}{*}{Scheduler} & \multicolumn{3}{c}{MC64} & \multicolumn{3}{c}{MC1} \\
\cmidrule(lr){3-5}\cmidrule(lr){6-8}
 &  & Train tok/s ($\times$ sync) & Overlap & Avg@32 & Train tok/s ($\times$ sync) & Overlap & Avg@32 \\
\midrule
\multirow{3}{*}{Qwen3-1.7B-Base} & Strict sync & 8.7k (1.00$\times$) & 0.81 & 8.85 & 8.6k (1.00$\times$) & 0.82 & \textbf{8.65} \\
 & Two-step-off & 14.2k (1.64$\times$) & 1.49 & 8.23 & 18.5k (2.15$\times$) & 1.65 & 8.12 \\
 & AsyncOPD (ours) & \textbf{23.4k (2.70$\times$)} & \textbf{2.13} & \textbf{9.38} & \textbf{28.1k (3.28$\times$)} & \textbf{2.31} & 8.44 \\
\midrule
\multirow{3}{*}{Qwen3-4B-Base} & Strict sync & 9.5k (1.00$\times$) & 0.81 & \textbf{25.00} & 8.1k (1.00$\times$) & 0.82 & 23.33 \\
 & Two-step-off & 12.4k (1.30$\times$) & 1.64 & 23.85 & 12.9k (1.60$\times$) & 1.65 & \textbf{23.54} \\
 & AsyncOPD (ours) & \textbf{15.8k (1.66$\times$)} & \textbf{2.19} & \textbf{25.00} & \textbf{16.4k (2.03$\times$)} & \textbf{2.27} & 23.44 \\
\midrule
\multirow{3}{*}{Qwen3-8B-Base} & Strict sync & 7.5k (1.00$\times$) & 0.81 & 26.56 & 6.5k (1.00$\times$) & 0.84 & 28.44 \\
 & Two-step-off & 11.6k (1.55$\times$) & 1.76 & 26.56 & 9.6k (1.47$\times$) & 1.63 & \textbf{28.85} \\
 & AsyncOPD (ours) & \textbf{14.5k (1.94$\times$)} & \textbf{2.24} & \textbf{28.65} & \textbf{10.6k (1.63$\times$)} & \textbf{2.17} & 27.50 \\
\bottomrule
\end{tabular}%
}
\end{table}

Train-time accuracy curves are reported in
\cref{app:asyncopd-scheduler-details}.

\section{Conclusion}
We present the first systematic study of staleness in asynchronous on-policy
distillation (OPD). Our results show that KL direction shapes the stale-data
problem: forward KL remains robust to stale rollouts, whereas reverse KL is more
vulnerable because it is student-weighted. In reverse-KL OPD, the most effective
policy-gradient surrogate uses the current advantage recomputed at learner time
without clipping; advanced asynchronous RL surrogates do not improve over this
choice. We also find that stale student top-$k$ supports are support-mismatched,
whereas one-sample Monte Carlo (MC) remains correctable but high-variance. This
contrast motivates multi-sample MC, which preserves MC correctability while
reducing one-sample variance. Finally, we present and open-source
\textbf{AsyncOPD}, a fully asynchronous OPD training pipeline built from these
estimator choices, improving training efficiency while maintaining OPD quality.

\paragraph{Limitations and Future Work}

We study sparse and Monte Carlo OPD estimators, not dense full-vocabulary KL in the asynchronous setting. Although dense KL avoids cached-support mismatch, it is difficult to implement efficiently when rollout, teacher scoring, and learner updates are decoupled. KDFlow~\citep{kdflow} suggests one path by transmitting teacher hidden states and recomputing student logits, but only for synchronous OPD. Extending this approach to asynchronous OPD while handling stale rollouts and preserving throughput is an important future direction.
Our experiments are also limited to a single 8-GPU node by available resources, not by the pipeline itself; scaling to larger multi-node clusters remains future work.

\begin{ack}
This work was supported by Institute for Information \& communications Technology Promotion (IITP) grant funded by the Korea government (MSIT) (No. 04-26-03-0081, Energy-Efficient Training–Inference System Optimization for Reinforcement Learning-Based Post-Training). This work was also supported by the “Advanced GPU Utilization Support Program” funded by the Government of the Republic of Korea (Ministry of Science and ICT).
\end{ack}

\bibliographystyle{abbrvnat}
\bibliography{reference}

@inproceedings{
areal,
title={{AREAL}: A Large-Scale Asynchronous Reinforcement Learning System for Language Reasoning},
author={Wei Fu and Jiaxuan Gao and Xujie Shen and Chen Zhu and Zhiyu Mei and Chuyi He and Shusheng Xu and Guo Wei and Jun Mei and WANG JIASHU and Tongkai Yang and Binhang Yuan and Yi Wu},
booktitle={The Thirty-ninth Annual Conference on Neural Information Processing Systems},
year={2025},
url={https://openreview.net/forum?id=X9diEuva9R}
}

@article{tml-opd,
  author = {Kevin Lu and Thinking Machines Lab},
  title = {On-Policy Distillation},
  journal = {Thinking Machines Lab: Connectionism},
  year = {2025},
  note = {https://thinkingmachines.ai/blog/on-policy-distillation},
  doi = {10.64434/tml.20251026},
}

@inproceedings{
minillm,
title={Mini{LLM}: Knowledge Distillation of Large Language Models},
author={Yuxian Gu and Li Dong and Furu Wei and Minlie Huang},
booktitle={The Twelfth International Conference on Learning Representations},
year={2024},
url={https://openreview.net/forum?id=5h0qf7IBZZ}
}

@inproceedings{
gkd,
title={On-Policy Distillation of Language Models: Learning from Self-Generated Mistakes},
author={Rishabh Agarwal and Nino Vieillard and Yongchao Zhou and Piotr Stanczyk and Sabela Ramos Garea and Matthieu Geist and Olivier Bachem},
booktitle={The Twelfth International Conference on Learning Representations},
year={2024},
url={https://openreview.net/forum?id=3zKtaqxLhW}
}

@inproceedings{verl,
  title={Hybridflow: A flexible and efficient rlhf framework},
  author={Sheng, Guangming and Zhang, Chi and Ye, Zilingfeng and Wu, Xibin and Zhang, Wang and Zhang, Ru and Peng, Yanghua and Lin, Haibin and Wu, Chuan},
  booktitle={Proceedings of the Twentieth European Conference on Computer Systems},
  pages={1279--1297},
  year={2025}
}

@article{g-opd,
  title={Learning beyond Teacher: Generalized On-Policy Distillation with Reward Extrapolation},
  author={Yang, Wenkai and Liu, Weijie and Xie, Ruobing and Yang, Kai and Yang, Saiyong and Lin, Yankai},
  journal={arXiv preprint arXiv:2602.12125},
  year={2026}
}

@inproceedings{
scalerl,
title={The Art of Scaling Reinforcement Learning Compute for {LLM}s},
author={Fnu Devvrit and Lovish Madaan and Rishabh Tiwari and Rachit Bansal and Sai Surya Duvvuri and Manzil Zaheer and Inderjit S Dhillon and David Brandfonbrener and Rishabh Agarwal},
booktitle={The Fourteenth International Conference on Learning Representations},
year={2026},
url={https://openreview.net/forum?id=FMjeC9Msws}
}

@inproceedings{
m2po,
title={Prosperity before Collapse: How Far Can Off-Policy {RL} Reach with Stale Data on {LLM}s?},
author={Haizhong Zheng and Jiawei Zhao and Beidi Chen},
booktitle={The Fourteenth International Conference on Learning Representations},
year={2026},
url={https://openreview.net/forum?id=IIgl5MWelz}
}

@article{areal-hex,
  title={AReaL-Hex: Accommodating Asynchronous RL Training over Heterogeneous GPUs},
  author={Yan, Ran and Jiang, Youhe and Wu, Tianyuan and Gao, Jiaxuan and Mei, Zhiyu and Fu, Wei and Mai, Haohui and Wang, Wei and Wu, Yi and Yuan, Binhang},
  journal={arXiv preprint arXiv:2511.00796},
  year={2025}
}

@inproceedings{
async-rlhf,
title={Faster, More Efficient {RLHF} through Off-Policy Asynchronous Learning},
author={Michael Noukhovitch and Shengyi Huang and Sophie Xhonneux and Arian Hosseini and Rishabh Agarwal and Aaron Courville},
booktitle={The Thirteenth International Conference on Learning Representations},
year={2025},
url={https://openreview.net/forum?id=FhTAG591Ve}
}

@article{qwen3,
  title={Qwen3 technical report},
  author={Yang, An and Li, Anfeng and Yang, Baosong and Zhang, Beichen and Hui, Binyuan and Zheng, Bo and Yu, Bowen and Gao, Chang and Huang, Chengen and Lv, Chenxu and others},
  journal={arXiv preprint arXiv:2505.09388},
  year={2025}
}

@article{li2026rethinking,
  title={Rethinking On-Policy Distillation of Large Language Models: Phenomenology, Mechanism, and Recipe},
  author={Li, Yaxuan and Zuo, Yuxin and He, Bingxiang and Zhang, Jinqian and Xiao, Chaojun and Qian, Cheng and Yu, Tianyu and Gao, Huan-ang and Yang, Wenkai and Liu, Zhiyuan and others},
  journal={arXiv preprint arXiv:2604.13016},
  year={2026}
}

@inproceedings{pytorch,
 author = {Paszke, Adam and Gross, Sam and Chintala, Soumith and Chanan, Gregory and Yang, Edward and DeVito, Zachary and Lin, Zeming and Desmaison, Alban and Antiga, Luca and Lerer, Adam},
 booktitle = {NIPS-W},
 title = {Automatic Differentiation in {PyTorch}},
 year = {2017}
}

@inproceedings{vllm,
 author = {Woosuk Kwon and Zhuohan Li and Siyuan Zhuang and Ying Sheng and Lianmin Zheng and Cody Hao Yu and Joseph E. Gonzalez and Hao Zhang and Ion Stoica},
 booktitle = {Proceedings of the ACM SIGOPS 29th Symposium on Operating Systems Principles},
 title = {Efficient Memory Management for Large Language Model Serving with {PagedAttention}},
 year = {2023}
}

@article{he2025deepmath,
  title={Deepmath-103k: A large-scale, challenging, decontaminated, and verifiable mathematical dataset for advancing reasoning},
  author={He, Zhiwei and Liang, Tian and Xu, Jiahao and Liu, Qiuzhi and Chen, Xingyu and Wang, Yue and Song, Linfeng and Yu, Dian and Liang, Zhenwen and Wang, Wenxuan and others},
  journal={arXiv preprint arXiv:2504.11456},
  year={2025}
}

@misc{aime2024,
  title        = {{AIME 2024}},
  year         = {2024},
  author={Zhang, Yifan and Math-AI, Team},
  howpublished = {\url{https://huggingface.co/datasets/Maxwell-Jia/AIME_2024}},
  note         = {Hugging Face dataset; accessed 2026-04-03}
}

@misc{aime2025,
  title        = {{AIME 2025}},
  year         = {2025},
  author={Zhang, Yifan and Math-AI, Team},
  howpublished = {\url{https://huggingface.co/datasets/yentinglin/aime_2025}},
  note         = {Hugging Face dataset; accessed 2026-04-03}
}

@misc{amc,
  author       = {{Mathematical Association of America}},
  title        = {{American Mathematics Competitions -- AMC}},
  year         = {2023},
  howpublished = {\url{https://maa.org/}},
  note         = {Accessed 2026-04-03}
}

@article{opd-survey,
  title={A Survey of On-Policy Distillation for Large Language Models},
  author={Song, Mingyang and Zheng, Mao},
  journal={arXiv preprint arXiv:2604.00626},
  year={2026}
}

@article{tip,
  title={TIP: Token Importance in On-Policy Distillation},
  author={Xu, Yuanda and Sang, Hejian and Zhou, Zhengze and He, Ran and Wang, Zhipeng and Geramifard, Alborz},
  journal={arXiv preprint arXiv:2604.14084},
  year={2026}
}

@article{streamrl,
  title={Streamrl: Scalable, heterogeneous, and elastic rl for llms with disaggregated stream generation},
  author={Zhong, Yinmin and Zhang, Zili and Song, Xiaoniu and Hu, Hanpeng and Jin, Chao and Wu, Bingyang and Chen, Nuo and Chen, Yukun and Zhou, Yu and Wan, Changyi and others},
  journal={arXiv preprint arXiv:2504.15930},
  year={2025}
}

@article{laminar,
  title={Laminar: A scalable asynchronous rl post-training framework},
  author={Sheng, Guangming and Tong, Yuxuan and Wan, Borui and Zhang, Wang and Jia, Chaobo and Wu, Xibin and Wu, Yuqi and Li, Xiang and Zhang, Chi and Peng, Yanghua and others},
  journal={arXiv preprint arXiv:2510.12633},
  year={2025}
}

@article{a-3po,
  title={A-3PO: Accelerating Asynchronous LLM Training with Staleness-aware Proximal Policy Approximation},
  author={Li, Xiaocan and Wu, Shiliang and Shen, Zheng},
  journal={arXiv preprint arXiv:2512.06547},
  year={2025}
}

@article{kdflow,
  title={KDFlow: A User-Friendly and Efficient Knowledge Distillation Framework for Large Language Models},
  author={Zhang, Songming and Zhang, Xue and Zhang, Tong and Hu, Bojie and Chen, Yufeng and Xu, Jinan},
  journal={arXiv preprint arXiv:2603.01875},
  year={2026}
}

@misc{deepseekv4,
      title={DeepSeek-V4: Towards Highly Efficient Million-Token Context Intelligence},
      author={DeepSeek-AI},
      year={2026},
}

@article{grpo,
  title={Deepseekmath: Pushing the limits of mathematical reasoning in open language models},
  author={Shao, Zhihong and Wang, Peiyi and Zhu, Qihao and Xu, Runxin and Song, Junxiao and Bi, Xiao and Zhang, Haowei and Zhang, Mingchuan and Li, YK and Wu, Yang and others},
  journal={arXiv preprint arXiv:2402.03300},
  year={2024}
}

@article{dapo,
  title={Dapo: An open-source llm reinforcement learning system at scale},
  author={Yu, Qiying and Zhang, Zheng and Zhu, Ruofei and Yuan, Yufeng and Zuo, Xiaochen and Yue, Yu and Dai, Weinan and Fan, Tiantian and Liu, Gaohong and Liu, Lingjun and others},
  journal={arXiv preprint arXiv:2503.14476},
  year={2025}
}

@article{deepseek-r1,
  title={Deepseek-r1: Incentivizing reasoning capability in llms via reinforcement learning},
  author={Guo, Daya and Yang, Dejian and Zhang, Haowei and Song, Junxiao and Wang, Peiyi and Zhu, Qihao and Xu, Runxin and Zhang, Ruoyu and Ma, Shirong and Bi, Xiao and others},
  journal={arXiv preprint arXiv:2501.12948},
  year={2025}
}

@article{rl-survey,
  title={A survey of reinforcement learning for large reasoning models},
  author={Zhang, Kaiyan and Zuo, Yuxin and He, Bingxiang and Sun, Youbang and Liu, Runze and Jiang, Che and Fan, Yuchen and Tian, Kai and Jia, Guoli and Li, Pengfei and others},
  journal={arXiv preprint arXiv:2509.08827},
  year={2025}
}

@article{opsd,
  title={Self-Distilled Reasoner: On-Policy Self-Distillation for Large Language Models},
  author={Zhao, Siyan and Xie, Zhihui and Liu, Mengchen and Huang, Jing and Pang, Guan and Chen, Feiyu and Grover, Aditya},
  journal={arXiv preprint arXiv:2601.18734},
  year={2026}
}

@article{glm-5,
  title={Glm-5: from vibe coding to agentic engineering},
  author={Zeng, Aohan and Lv, Xin and Hou, Zhenyu and Du, Zhengxiao and Zheng, Qinkai and Chen, Bin and Yin, Da and Ge, Chendi and Huang, Chenghua and Xie, Chengxing and others},
  journal={arXiv preprint arXiv:2602.15763},
  year={2026}
}

@article{mimo-v2-flash,
  title={Mimo-v2-flash technical report},
  author={Xiao, Bangjun and Xia, Bingquan and Yang, Bo and Gao, Bofei and Shen, Bowen and Zhang, Chen and He, Chenhong and Lou, Chiheng and Luo, Fuli and Wang, Gang and others},
  journal={arXiv preprint arXiv:2601.02780},
  year={2026}
}

@article{rollpacker,
  title={Rollpacker: Mitigating long-tail rollouts for fast, synchronous rl post-training},
  author={Gao, Wei and Zhao, Yuheng and An, Dakai and Wu, Tianyuan and Cao, Lunxi and Xiong, Shaopan and Huang, Ju and Wang, Weixun and Yang, Siran and Su, Wenbo and others},
  journal={arXiv preprint arXiv:2509.21009},
  year={2025}
}

@article{ppo,
  title={Proximal policy optimization algorithms},
  author={Schulman, John and Wolski, Filip and Dhariwal, Prafulla and Radford, Alec and Klimov, Oleg},
  journal={arXiv preprint arXiv:1707.06347},
  year={2017}
}

\appendix

\section{Sparse and Monte Carlo Reverse-KL Implementations}
\label{app:sparse-implementations}

\subsection{Sparse Top-\texorpdfstring{$k$}{k} Reverse-KL OPD}
\label{app:sparse-topk}

The dense reverse-KL objective in \cref{eq:dense-reverse-kl} sums over the full
vocabulary. A sparse top-$k$ implementation instead evaluates reverse KL on a
finite student support
\begin{align}
S_\theta(s)=\operatorname{TopK}\!\left(p_\theta(\cdot\mid s),k\right).
\end{align}
For any support $S$, define the restricted normalizers
$Z_p^S(s)=\sum\nolimits_{u\in S}p_\theta(u\mid s)$ and
$Z_q^S(s)=\sum\nolimits_{u\in S}q(u\mid s)$, and the renormalized
distributions
\begin{align}
\tilde p_\theta^S(a\mid s)
=
\frac{p_\theta(a\mid s)\mathbf{1}[a\in S]}{Z_p^S(s)},
\qquad
\tilde q^S(a\mid s)
=
\frac{q(a\mid s)\mathbf{1}[a\in S]}{Z_q^S(s)}.
\end{align}
The sparse reverse-KL objective is
\begin{align}
D_R^S(\theta;s)
&=
\mathrm{KL}\!\left(\tilde p_\theta^S(\cdot\mid s)\,\|\,\tilde q^S(\cdot\mid s)\right)
\nonumber\\
&=
-\sum\nolimits_{a\in S}
\tilde p_\theta^S(a\mid s)
\left(
\log \tilde q^S(a\mid s)-\log \tilde p_\theta^S(a\mid s)
\right).
\end{align}
In practice, when $S=S_\theta(s)$, we treat the selected top-$k$ support as
fixed during the local update.

\subsection{Monte Carlo Reverse-KL OPD}
\label{app:mc-reverse-kl}

Let $A_\theta(a,s)=\log q(a\mid s)-\log p_\theta(a\mid s)$. From
\cref{eq:reverse-kl-gradient}, the dense reverse-KL gradient can be written as
\begin{align}
\nabla_\theta D_R(\theta;s)
=
-\mathbb{E}_{a\sim p_\theta(\cdot\mid s)}
\left[
A_\theta(a,s)\nabla_\theta\log p_\theta(a\mid s)
\right].
\end{align}
A one-sample current-policy Monte Carlo estimator is therefore
\begin{align}
\widehat g_{\mathrm{MC}}(s,a)
=
-A_\theta(a,s)\nabla_\theta\log p_\theta(a\mid s),
\qquad
a\sim p_\theta(\cdot\mid s).
\end{align}
With $m$ independent samples $a_i\sim p_\theta(\cdot\mid s)$, the corresponding
multi-sample estimator averages the same local term:
\begin{align}
\widehat g_m(s)
=
-\frac{1}{m}\sum\nolimits_{i=1}^{m}
A_\theta(a_i,s)\nabla_\theta\log p_\theta(a_i\mid s).
\end{align}

\section{Experimental Details}
\label{app:experimental-details}

This section details the experimental setup. Unless explicitly stated otherwise, experiments use the common setup in \cref{tab:app-common-experimental-settings} and report final-checkpoint Avg@32 accuracy. Our implementation uses vLLM~\citep{vllm} for rollout generation and teacher scoring, PyTorch FSDP~\citep{pytorch} for learner training, and runs each experiment on a single 8$\times$B200 node. Individual experiments take roughly 1--12 hours, depending on the setting. Asset URLs, license names, and versions are summarized in \cref{tab:app-asset-licenses}.

\begin{table}[t]
\centering
\small
\setlength{\tabcolsep}{5pt}
\renewcommand{\arraystretch}{1.08}
\caption{Experimental settings.}
\label{tab:app-common-experimental-settings}
\begin{tabular}{@{}p{0.26\textwidth}p{0.68\textwidth}@{}}
\toprule
Setting & Value \\
\midrule
Student & Qwen3-4B-Base for main staleness/estimator experiments; AsyncOPD experiments also use Qwen3-\{1.7B,4B,8B\}-Base and thinking-disabled Qwen3-\{1.7B,4B,8B\}~\citep{qwen3}. \\
Teacher & Qwen3-30B-A3B-Instruct-2507~\citep{qwen3} \\
Training data & DeepMath~\citep{he2025deepmath} filtered to 57{,}630 math problems with difficulty level at least 6 \\
Prompt / response lengths & 2{,}048 prompt tokens / 16{,}384 response tokens \\
Training horizon & 200 training iterations; in the common setup each iteration contains four mini-batch optimizer updates \\
Optimization & batch size 256, mini-batch size 64 \\
Optimizer & AdamW-style optimizer with learning rate $3\times 10^{-6}$, constant schedule, weight decay 0.01 \\
PPO clipping & $\epsilon=0.2$ for clipped PPO-style surrogates \\
M2PO budget & $0.01$ \\
Rollout generation & temperature 1.0, top-$p=1.0$, no top-$k$ truncation \\
Evaluation generation & 32 samples per problem, temperature 1.0 \\
Implementation & vLLM rollout and teacher scoring; PyTorch FSDP learner training \\
Hardware & Single 8$\times$B200 node \\
GPU allocation & One teacher GPU; strict sync time-shares seven GPUs between rollout and training; stale-cache sweeps and AsyncOPD use concurrent rollout and trainer GPU pools \\
\bottomrule
\end{tabular}
\end{table}

\paragraph{Constructing the staleness axis.}
The main text uses staleness as an experimental control over how old the cached
rollout data is when the learner updates on it. In all staleness plots and
tables in \cref{sec:opd-staleness,sec:reverse-kl-estimator-design,sec:estimator-design},
staleness is measured in train-batch steps. One train-batch step is one logical
rollout batch consumed by the learner for a training iteration. The sweep value
$k$ is therefore the target number of train-batch steps by which the consumed
cache trails the current learner; equivalently, it is the target cache depth in
logical rollout batches. A value $k=0$ is synchronous: rollout, teacher scoring,
training, and weight synchronization occur in strict sequence. For $k>0$, the
run first generates exactly $k$ rollout batches with the initial student
snapshot before the first learner update. Training then consumes the oldest
available generated batch; after each learner update and weight synchronization,
a new rollout batch is generated with the latest student snapshot whenever
needed to restore the target cache depth.

This protocol is the operational source of the prefix- and action-level
staleness discussed in the main text: the consumed prefixes and cached actions
come from an older rollout student, while the update is applied to the current
student. For a consumed rollout batch, let $t_{\mathrm{roll}}$ be the
train-batch index of the student snapshot used for generation and
$t_{\mathrm{train}}$ be the train-batch index at learner time. The staleness
used in the plots is
\[
\Delta_{\mathrm{batch}} = t_{\mathrm{train}} - t_{\mathrm{roll}} .
\]
All examples in a logical batch share the same rollout snapshot and therefore
share the same $\Delta_{\mathrm{batch}}$. Under the controlled cache protocol,
$\Delta_{\mathrm{batch}}$ ramps as $0,1,2,\ldots$ while the initial cache is
drained and then plateaus at $k$. Thus a 64-batch target cache depth is plotted
as staleness 64. A train-batch step can contain multiple mini-batch optimizer
updates; in the common setup, $B=256$ and $B_{\mathrm{mini}}=64$, so each
train-batch step contains $M=B/B_{\mathrm{mini}}=4$ optimizer updates. This
conversion is useful for implementation accounting, but it is not the staleness
axis used in the plots.

We use $k$ as the x-axis because it is the controlled train-batch staleness
intervention shared across methods. The sweep covers
$k\in\{0,1,2,4,8,16,32,64,128\}$ across the forward-KL, reverse-KL /
PPO-style, M2PO / DecPPO, top-$k$, and Monte-Carlo support-size variants; apart
from the estimator choice and $k$, these runs share the common model, data,
batch-size, generation, and evaluation settings in
\cref{tab:app-common-experimental-settings}.

\section{Datasets and Metrics}
\label{app:datasets-metrics}

\paragraph{Training data.}
We filter the DeepMath dataset~\citep{he2025deepmath} to retain 57{,}630 math problems with difficulty level greater than or equal to 6, and use this filtered subset as the training data.

\paragraph{Evaluation datasets.}
\Cref{tab:app-eval-datasets} lists the evaluation datasets used: AIME 2024~\citep{aime2024}, AIME 2025~\citep{aime2025}, and AMC 2023~\citep{amc}. AIME24 is evaluated every 20 steps. The remaining datasets are only evaluated for the final checkpoint.

\begin{table}[t]
\centering
\small
\setlength{\tabcolsep}{5pt}
\renewcommand{\arraystretch}{1.08}
\caption{Evaluation datasets.}
\label{tab:app-eval-datasets}
\begin{tabular}{@{}lr@{}}
\toprule
Dataset & Problems \\
\midrule
AIME 2024 & 30 \\
AIME 2025 & 30 \\
AMC 2023 & 40 \\
\bottomrule
\end{tabular}
\end{table}

\paragraph{Accuracy metric.}
Evaluation samples 32 responses per problem. For a dataset $D$, the reported Avg@32 is the mean per-problem pass rate,
\begin{equation}
\mathrm{Avg@32}(D)=100\cdot\frac{1}{|D|}\sum_{i\in D}\frac{c_i}{32},
\end{equation}
where $c_i$ is the number of sampled responses judged correct for problem $i$. Paper tables and plots use Avg@32 unless noted otherwise.

\section{Existing Asset Licenses}
\label{app:asset-licenses}

\Cref{tab:app-asset-licenses} lists the reused assets.

\begin{table}[t]
\centering
\scriptsize
\setlength{\tabcolsep}{2pt}
\renewcommand{\arraystretch}{1.15}
\caption{Existing assets used in this work, with source URLs, license names, and versions.}
\label{tab:app-asset-licenses}
\resizebox{\linewidth}{!}{%
\begin{tabular}{@{}p{0.10\textwidth}p{0.19\textwidth}p{0.31\textwidth}p{0.20\textwidth}p{0.13\textwidth}@{}}
\toprule
Category & Asset & URL & License name & Version \\
\midrule
Dataset & DeepMath & \url{https://huggingface.co/datasets/zwhe99/DeepMath-103K} & MIT & 2025 \\
Dataset & AIME 2024 & \url{https://huggingface.co/datasets/Maxwell-Jia/AIME_2024} & MIT & 2024 \\
Dataset & AIME 2025 & \url{https://huggingface.co/datasets/yentinglin/aime_2025} & Not specified & 2025 \\
Dataset & AMC 2023 / American Mathematics Competitions & \url{https://huggingface.co/datasets/math-ai/amc23} & Not specified & 2023 \\
Model & Qwen3-4B-Base & \url{https://huggingface.co/Qwen/Qwen3-4B-Base} & Apache-2.0 & Not specified \\
Model & Qwen3-1.7B-Base & \url{https://huggingface.co/Qwen/Qwen3-1.7B-Base} & Apache-2.0 & Not specified \\
Model & Qwen3-8B-Base & \url{https://huggingface.co/Qwen/Qwen3-8B-Base} & Apache-2.0 & Not specified \\
Model & Qwen3-4B & \url{https://huggingface.co/Qwen/Qwen3-4B} & Apache-2.0 & Not specified \\
Model & Qwen3-1.7B & \url{https://huggingface.co/Qwen/Qwen3-1.7B} & Apache-2.0 & Not specified \\
Model & Qwen3-8B & \url{https://huggingface.co/Qwen/Qwen3-8B} & Apache-2.0 & Not specified \\
Model & Qwen3-30B-A3B-Instruct-2507 & \url{https://huggingface.co/Qwen/Qwen3-30B-A3B-Instruct-2507} & Apache-2.0 & Qwen3-2507 \\
Software & vLLM & \url{https://github.com/vllm-project/vllm} & Apache-2.0 & 0.16.0 \\
Software & PyTorch / PyTorch FSDP & \url{https://github.com/pytorch/pytorch} & BSD-3-Clause & 2.9.1 \\
\bottomrule
\end{tabular}%
}
\end{table}

\section{Multi-Sample MC Variance at Large Staleness}
\label{app:mc-variance-large-staleness}

We measure how multi-sample MC reduces the variance of the old-to-current IS
surrogate at large staleness.  At timestep $t$ with prefix $s_t$, local MC
actions $a_{t,1},\ldots,a_{t,m}$ are sampled iid with replacement from
$p_{\mathrm{old}}(\cdot\mid s_t)$ (duplicates are allowed), and the learner
evaluates
\begin{align}
\widehat L_m^{\mathrm{MC}}(\theta;s_t)
=
-\frac{1}{m}\sum_{i=1}^{m}
\rho_\theta(a_{t,i},s_t)
\operatorname{sg}\!\left(A_\theta(a_{t,i},s_t)\right),
\qquad
\rho_\theta(a,s)=\frac{p_\theta(a\mid s)}{p_{\mathrm{old}}(a\mid s)}.
\end{align}
Using the Qwen3-4B-Base staleness-128 runs, we report $R_m^{\mathrm{local}}$
in the fixed-prefix column and $R_m^{\mathrm{seq}}$ in the sequence-level
column of \cref{tab:app-mc-variance-large-staleness}.  Both ratios are
normalized to the corresponding $m=1$ estimator within the same old-to-current
pair:
\begin{align}
R_m^{\mathrm{local}}
&=
\frac{
\mathbb{E}_{s_t}\!\left[
\operatorname{Var}_{a_{t,1},\ldots,a_{t,m}
\sim p_{\mathrm{old}}(\cdot\mid s_t)}
\left(\widehat L_m^{\mathrm{MC}}(\theta;s_t)\mid s_t\right)
\right]}
{
\mathbb{E}_{s_t}\!\left[
\operatorname{Var}_{a_t\sim p_{\mathrm{old}}(\cdot\mid s_t)}
\left(\widehat L_1^{\mathrm{MC}}(\theta;s_t)\mid s_t\right)
\right]}, \\
R_m^{\mathrm{seq}}
&=
\frac{
\operatorname{Var}\!\left[
\frac{1}{T}\sum_{t=1}^{T}\widehat L_m^{\mathrm{MC}}(\theta;s_t)
\right]}
{
\operatorname{Var}\!\left[
\frac{1}{T}\sum_{t=1}^{T}\widehat L_1^{\mathrm{MC}}(\theta;s_t)
\right]} .
\end{align}
For $R_m^{\mathrm{seq}}$, the generated prefix path $s_{1:T}$ is fixed when
computing the variance; the MC samples are local scorer queries at each fixed
prefix, not separate rollout branches.

\begin{table}[!ht]
\centering
\scriptsize
\setlength{\tabcolsep}{6pt}
\renewcommand{\arraystretch}{1.08}
\caption{MC variance ratios at large staleness.  The fixed-prefix column
isolates local next-token action-sampling variance; the sequence-level column
averages the same estimator over generated timesteps before computing
variance.}
\label{tab:app-mc-variance-large-staleness}
\begin{tabular}{@{}rrr@{}}
\toprule
Diagnostic $m$ & Fixed-prefix ratio $R_m^{\mathrm{local}}$ &
Sequence-level ratio $R_m^{\mathrm{seq}}$ \\
\midrule
1  & 1.000 & 1.000 \\
4  & 0.255 & 0.338 \\
16 & 0.0588 & 0.152 \\
64 & 0.0149 & 0.112 \\
\bottomrule
\end{tabular}
\end{table}

\Cref{tab:app-mc-variance-large-staleness} shows that larger $m$ consistently
reduces variance.  Fixed-prefix ratios closely follow the $1/m$ reference
($m=64$ leaves $1.49\%$ of the one-sample local variance, versus
$1/64=1.56\%$).  After timestep aggregation, $m=64$ still leaves only
$11.2\%$ of the one-sample sequence-level variance, showing that multi-sample
MC reduces variance in practice even though sequence-level averaging makes the
reduction less extreme than the local next-token effect.

\section{Importance-Sampling Ablation}
\label{app:is-ablation}

\begin{figure}[!ht]
\centering
\begin{subfigure}[t]{0.24\linewidth}
\centering
\includegraphics[width=\linewidth]{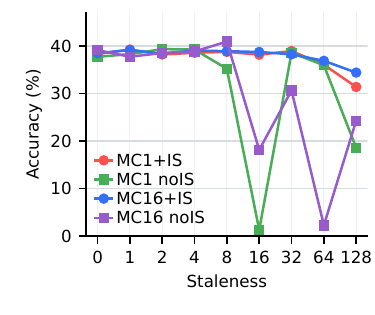}
\caption{Average}
\label{fig:mean-mc-is-ablation-final}
\end{subfigure}
\hfill
\begin{subfigure}[t]{0.24\linewidth}
\centering
\includegraphics[width=\linewidth]{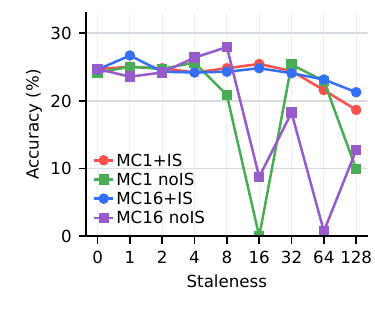}
\caption{AIME24}
\label{fig:aime24-mc-is-ablation-final}
\end{subfigure}
\hfill
\begin{subfigure}[t]{0.24\linewidth}
\centering
\includegraphics[width=\linewidth]{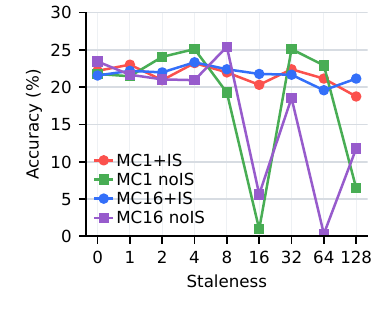}
\caption{AIME25}
\label{fig:aime25-mc-is-ablation-final}
\end{subfigure}
\hfill
\begin{subfigure}[t]{0.24\linewidth}
\centering
\includegraphics[width=\linewidth]{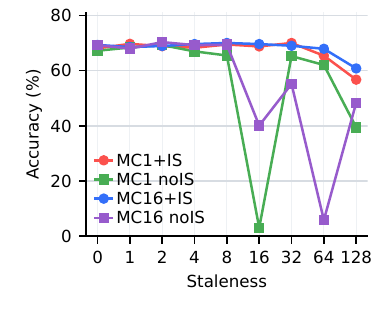}
\caption{AMC}
\label{fig:amc-mc-is-ablation-final}
\end{subfigure}
\caption{Accuracy comparison under staleness for MC importance-sampling ablations. Increasing the number of samples reduces Monte Carlo variance, but old-to-current IS is still needed to correct stale-policy sampling.}
\label{fig:aime24-mc-is-ablation}
\end{figure}

Before treating multi-sampling as an improvement, we separate it from IS.
Increasing $m$ changes the Monte Carlo variance of the estimator. It does not
define the target distribution. Old-to-current IS is still the mechanism that
turns behavior-policy samples into an estimator of the current reverse-KL local
gradient.
\Cref{fig:aime24-mc-is-ablation} compares MC1 and MC16 with and without IS to
test this distinction directly.

\section{AsyncOPD Scheduler Details}
\label{app:asyncopd-scheduler-details}

This appendix gives the implementation details omitted from \cref{sec:asyncopd}.
Our scheduler, AsyncOPD, follows the fully asynchronous systems structure of
AReaL~\citep{areal}, but the queue contains OPD cache items rather than
reward-labeled RL trajectories.

\paragraph{Queue interface.}
The pipeline has three long-running stages: rollout generation, teacher scoring,
and learner training. Rollout workers sample trajectories from their latest
synchronized student snapshot. For each visited prefix $s$, they cache the MC
actions, rollout log probabilities under $p_{\mathrm{old}}$, and the rollout
student version. The main scheduler comparison uses MC64; the MC1 runs use the
same queue interface with one cached action. The teacher scores the cached
actions. The learner then recomputes $\log p_\theta(a\mid s)$ and
$A_\theta(a,s)$ under the current student, and applies the unclipped
old-to-current IS estimator from \cref{sec:estimator-design}.

\paragraph{Weight synchronization and queue capacity.}
During AsyncOPD weight synchronization, rollout workers pause generation. In the
keep-mode used for the scheduler experiments, in-flight requests are not
discarded: the already sampled token prefix is kept, the student weights are
updated, and the running-request prefix cache is reset so the engine rebuilds
the attention state for that prefix under the new weights before generation
resumes. Thus, tokens before the synchronization point are reused rather than
regenerated, while later tokens are sampled under the new student snapshot. Each
completed sample records the token index at which the active weight version
changes. The queue-depth parameter $\tau$ is enforced as a capacity bound rather
than as a learner-side drop rule. The coordinator creates a semaphore with
$(\tau+1)B$ permits, where $B$ is the effective train batch size. The prompt
feeder acquires one permit before submitting a prompt to rollout, and the train
dispatcher releases permits only after the corresponding samples have been
consumed by a learner update. During weight synchronization, a sync gate
prevents the feeder from using newly released permits until rollout workers have
received the updated weights. Thus, smaller $\tau$ limits the amount of
unconsumed rollout work in the pipeline, while larger $\tau$ permits a deeper
backlog and more overlap. The queues themselves remain FIFO; items are not
evicted for being stale.

\paragraph{Training throughput metric.}
The table reports training throughput. Let $n_j$ be the number of
response tokens used by learner update $j$, and let $t_j$ be the train
wall-clock time after that update. Discarding the first five warmup updates, we
compute
\[
  \mathrm{throughput}
  = \frac{\sum_{j=6}^{J} n_j}{\sum_{j=6}^{J}(t_j-t_{j-1})}
  = \frac{\sum_{j=6}^{J} n_j}{t_J - t_5}.
\]
Speedups are normalized to the strict-sync run with the same student and MC
setting.

\paragraph{Pipeline overlap metric.}
Let $\mathcal{S}=\{\text{rollout},\text{teacher},\text{train}\}$.
For teacher and train, we merge overlapping busy intervals within each stage and
compute the merged busy time $T_s$. Rollout has $N_r$ workers, so we
first merge intervals within each worker $i$ and then define the rollout-stage
busy time as the worker-normalized average
\[
  T_{\mathrm{rollout}} = \frac{1}{N_r}\sum_{i=1}^{N_r} T_{\mathrm{rollout},i}.
\]
Let $T_{\mathrm{wall}}$ be the elapsed train wall-clock
interval from the first to the last recorded pipeline-stage interval. We define
\[
  \mathrm{overlap} = \frac{\sum_{s\in\mathcal{S}} T_s}{T_{\mathrm{wall}}}.
\]
A mostly serial schedule has overlap near $1$, and the maximum remains $3$:
all rollout workers, teacher scoring, and training busy for the full interval.

\paragraph{Hardware and testing protocol.}
Each scheduler run uses one 8-GPU node. One GPU is reserved for teacher scoring.
The remaining seven GPUs are the rollout/training pool. Rollout generation uses
data parallelism, and learner training uses PyTorch FSDP. Strict sync runs
time-share this pool: all seven GPUs run rollout, then all seven switch to
training, and the cycle repeats. The two-step-off and our AsyncOPD runs
split the same seven GPUs concurrently: 4 GPUs for rollout workers and 3 GPUs
for the FSDP trainer.

For each student size and MC setting, we compare strict sync, two-step-off, and
our AsyncOPD scheduler with the same teacher, training data, evaluation
metrics, and reverse-KL estimator: current-policy $A_\theta$, no clipping, and
old-to-current IS correction. Two-step-off fixes a two-update offset between
rollout and the learner update that consumes it, so stale rollout reuse is
static and controlled rather than produced by queue timing. We use this offset
because it is the fastest static step-off schedule under the 4-rollout/3-trainer
split. The OPD pipeline has three serial stages: rollout generation, teacher
scoring, and learner training. Therefore, a two-step offset is enough to keep all
stages occupied in the gated schedule. Larger offsets only make the consumed data
older; they do not create another OPD stage to overlap or remove the step-off
batch barrier. We measure final-checkpoint Avg@32 and train wall-clock time
over the same training horizon.

\begin{figure}[t]
\centering
\begin{subfigure}[t]{0.32\linewidth}
\centering
\includegraphics[width=\linewidth]{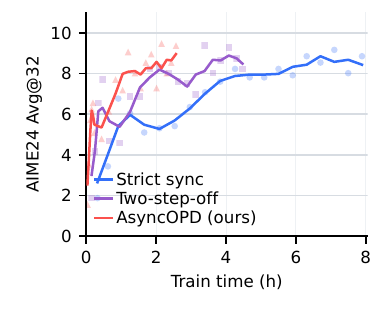}
\caption{MC64, 1.7B-Base}
\label{fig:asyncopd-accuracy-mc64-1-7b-base}
\end{subfigure}
\hfill
\begin{subfigure}[t]{0.32\linewidth}
\centering
\includegraphics[width=\linewidth]{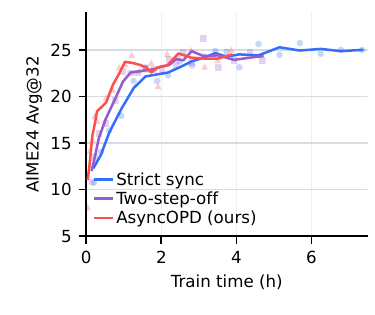}
\caption{MC64, 4B-Base}
\label{fig:asyncopd-accuracy-mc64-4b-base}
\end{subfigure}
\hfill
\begin{subfigure}[t]{0.32\linewidth}
\centering
\includegraphics[width=\linewidth]{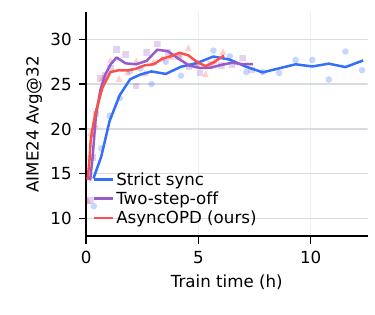}
\caption{MC64, 8B-Base}
\label{fig:asyncopd-accuracy-mc64-8b-base}
\end{subfigure}
\par\medskip
\begin{subfigure}[t]{0.32\linewidth}
\centering
\includegraphics[width=\linewidth]{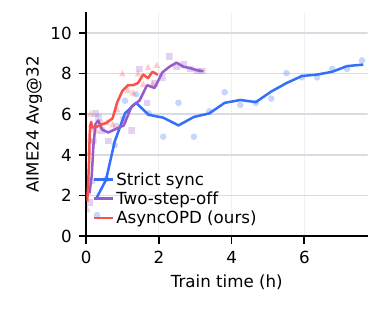}
\caption{MC1, 1.7B-Base}
\label{fig:asyncopd-accuracy-mc1-1-7b-base}
\end{subfigure}
\hfill
\begin{subfigure}[t]{0.32\linewidth}
\centering
\includegraphics[width=\linewidth]{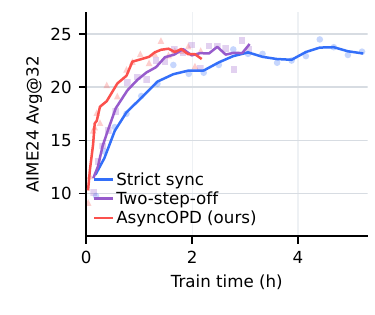}
\caption{MC1, 4B-Base}
\label{fig:asyncopd-accuracy-mc1-4b-base}
\end{subfigure}
\hfill
\begin{subfigure}[t]{0.32\linewidth}
\centering
\includegraphics[width=\linewidth]{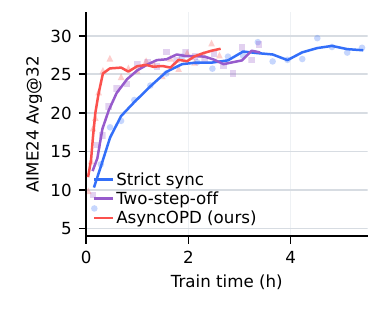}
\caption{MC1, 8B-Base}
\label{fig:asyncopd-accuracy-mc1-8b-base}
\end{subfigure}
\caption{Train-time AIME24 Avg@32 for Qwen3-Base students with MC64 and MC1. Lines are 3-point moving averages; faint markers are raw evaluations; colors denote scheduler. AsyncOPD reaches later checkpoints sooner, so its accuracy improves earlier in wall-clock time.}
\label{fig:asyncopd-accuracy-by-loss-model}
\end{figure}

\paragraph{Qwen3-Base train-time accuracy.}
\Cref{fig:asyncopd-accuracy-by-loss-model} provides the train-time view for
Qwen3-Base students. AsyncOPD reaches later checkpoints sooner, so accuracy
improves earlier in wall-clock time across student sizes and MC settings.

\begin{figure}[t]
\centering
\begin{subfigure}[t]{0.32\linewidth}
\centering
\includegraphics[width=\linewidth]{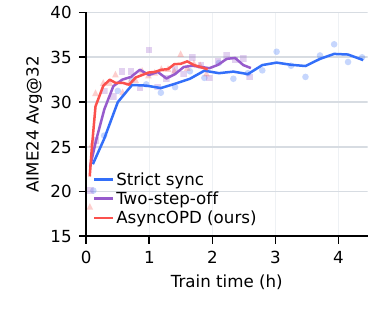}
\caption{MC64, 1.7B}
\label{fig:asyncopd-accuracy-mc64-1-7b}
\end{subfigure}
\hfill
\begin{subfigure}[t]{0.32\linewidth}
\centering
\includegraphics[width=\linewidth]{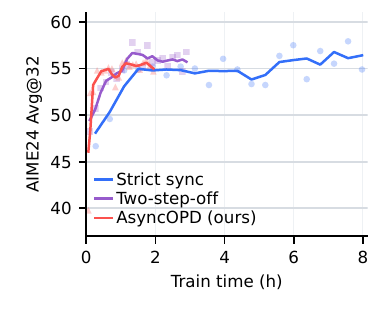}
\caption{MC64, 4B}
\label{fig:asyncopd-accuracy-mc64-4b}
\end{subfigure}
\hfill
\begin{subfigure}[t]{0.32\linewidth}
\centering
\includegraphics[width=\linewidth]{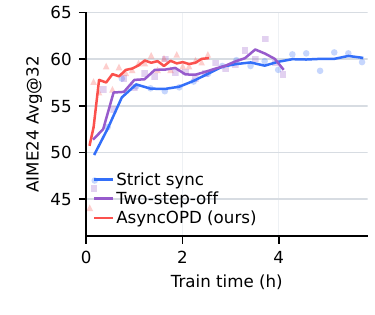}
\caption{MC64, 8B}
\label{fig:asyncopd-accuracy-mc64-8b}
\end{subfigure}
\par\medskip
\begin{subfigure}[t]{0.32\linewidth}
\centering
\includegraphics[width=\linewidth]{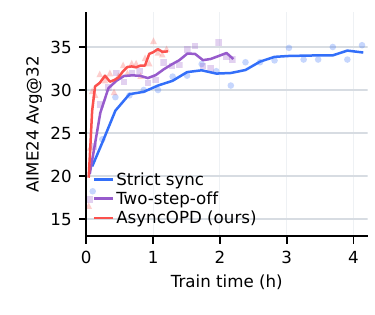}
\caption{MC1, 1.7B}
\label{fig:asyncopd-accuracy-mc1-1-7b}
\end{subfigure}
\hfill
\begin{subfigure}[t]{0.32\linewidth}
\centering
\includegraphics[width=\linewidth]{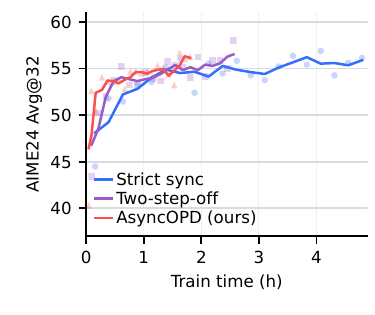}
\caption{MC1, 4B}
\label{fig:asyncopd-accuracy-mc1-4b}
\end{subfigure}
\hfill
\begin{subfigure}[t]{0.32\linewidth}
\centering
\includegraphics[width=\linewidth]{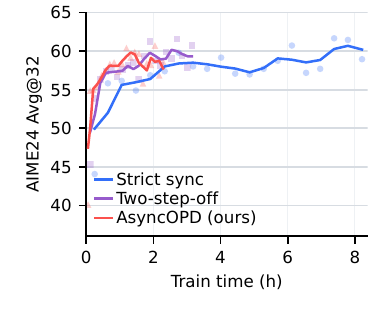}
\caption{MC1, 8B}
\label{fig:asyncopd-accuracy-mc1-8b}
\end{subfigure}
\caption{Train-time AIME24 Avg@32 for Qwen3 1.7B, 4B, and 8B students with thinking disabled, using MC64 and MC1. Lines are 3-point moving averages; faint markers are raw evaluations; colors denote scheduler. AsyncOPD reaches later checkpoints sooner, so its accuracy improves earlier in wall-clock time.}
\label{fig:asyncopd-accuracy-qwen3-by-loss-model}
\end{figure}

\paragraph{Additional Qwen3 AsyncOPD results.}
For the Qwen3 1.7B, 4B, and 8B student rows, we disable thinking at the
tokenizer prompt-formatting level: prompt construction uses the Qwen3
tokenizer's non-thinking chat-template mode before rollout and evaluation.
\Cref{tab:asyncopd-scheduler-qwen3-results,fig:asyncopd-accuracy-qwen3-by-loss-model}
report this comparison. The systems pattern matches the
Qwen3-Base results: AsyncOPD has the highest throughput and overlap, 
reaching up to $3.8\times$ strict-sync throughput on MC64 and up to
$3.2\times$ on MC1. The train-time accuracy plots show the same wall-clock
pattern as the main Qwen3-Base results: AsyncOPD reaches later checkpoints
sooner, so accuracy improves earlier across student sizes and MC settings.

\begin{table}[t]
\centering
\footnotesize
\setlength{\tabcolsep}{3pt}
\renewcommand{\arraystretch}{1.08}
\caption{Additional AsyncOPD scheduler results for Qwen3 students with thinking disabled. Train tok/s is training throughput; parentheses show speedup over the matched strict-sync baseline. Overlap is concurrent OPD-stage activity. Avg@32 is final AIME24. AsyncOPD achieves the highest throughput and overlap in all matched settings while maintaining comparable final accuracy.}
\label{tab:asyncopd-scheduler-qwen3-results}
\resizebox{\linewidth}{!}{%
\begin{tabular}{@{}llrrrrrr@{}}
\toprule
Student & Scheduler & \multicolumn{3}{c}{MC64} & \multicolumn{3}{c}{MC1} \\
\cmidrule(lr){3-5}\cmidrule(lr){6-8}
 &  & Train tok/s ($\times$ sync) & Overlap & Avg@32 & Train tok/s ($\times$ sync) & Overlap & Avg@32 \\
\midrule
\multirow{3}{*}{1.7B} & Strict sync & 11.2k (1.00$\times$) & 0.78 & \textbf{35.00} & 11.7k (1.00$\times$) & 0.78 & \textbf{35.21} \\
 & Two-step-off & 18.0k (1.61$\times$) & 1.58 & 32.81 & 20.8k (1.78$\times$) & 1.71 & 33.54 \\
 & AsyncOPD (ours) & \textbf{24.2k (2.16$\times$)} & \textbf{2.07} & 33.23 & \textbf{37.0k (3.17$\times$)} & \textbf{2.56} & 34.79 \\
\midrule
\multirow{3}{*}{4B} & Strict sync & 4.5k (1.00$\times$) & 0.87 & 54.90 & 7.4k (1.00$\times$) & 0.83 & 56.15 \\
 & Two-step-off & 11.5k (2.58$\times$) & 1.54 & \textbf{56.77} & 13.0k (1.75$\times$) & 1.63 & \textbf{58.02} \\
 & AsyncOPD (ours) & \textbf{17.0k (3.82$\times$)} & \textbf{2.15} & 54.69 & \textbf{18.5k (2.49$\times$)} & \textbf{2.26} & 56.25 \\
\midrule
\multirow{3}{*}{8B} & Strict sync & 5.6k (1.00$\times$) & 0.86 & 59.69 & 3.9k (1.00$\times$) & 0.87 & 58.96 \\
 & Two-step-off & 7.5k (1.34$\times$) & 1.45 & 58.33 & 9.6k (2.45$\times$) & 1.61 & \textbf{60.73} \\
 & AsyncOPD (ours) & \textbf{10.1k (1.81$\times$)} & \textbf{2.11} & \textbf{60.52} & \textbf{10.9k (2.78$\times$)} & \textbf{2.17} & 58.65 \\
\bottomrule
\end{tabular}%
}
\end{table}

\end{document}